\documentclass[sn-mathphys-num]{sn-jnl}% Math and Physical Sciences Numbered Reference Style 
\usepackage{multirow}%
\usepackage{amsmath,amssymb,amsfonts}%
\usepackage{amsthm}%
\usepackage{mathrsfs}%
\usepackage[title]{appendix}%
\usepackage{xcolor}%
\usepackage{textcomp}%
\usepackage{manyfoot}%
\usepackage{booktabs}%
\usepackage{algorithm}%
\usepackage{algorithmicx}%
\usepackage{algpseudocode}%
\usepackage{listings}%
\usepackage{svg}
\usepackage{subcaption}
\usepackage{makecell}
\usepackage{graphicx}
\usepackage{adjustbox}

\usepackage{pgfplots}
\usetikzlibrary{intersections}
\usepackage{bm}
\usetikzlibrary{positioning}
\usepgfplotslibrary{fillbetween}
\pgfplotsset{compat=1.5}
\usetikzlibrary{pgfplots.groupplots}
\usetikzlibrary{shapes.geometric,backgrounds}

\usepackage{tikz}
\usepackage{xifthen}

\pgfdeclarelayer{bg}    % declare background layer
\pgfsetlayers{bg,main}  % set the order of the layers (main is the standard layer)
\definecolor{beige}{RGB}{245, 245, 220}

\definecolor{darkgrey}{RGB}{75, 75, 75}
\definecolor{lightgrey}{RGB}{250, 250, 250}

\usetikzlibrary{calc, arrows, fit, positioning, patterns, decorations.pathreplacing, shapes}
\tikzstyle{dash} = [dashed, -latex,>=latex]
\tikzstyle{line} = [draw, -latex,>=latex]
\tikzstyle{smallbox} = [draw, minimum size=5.0mm]
\tikzstyle{box} = [draw, minimum size=7.0mm]
\tikzstyle{bigbox} = [draw, minimum size=10.0mm]
\tikzstyle{rectangle} = [draw, minimum width=10.0mm, minimum height=20.0mm]
\tikzstyle{switch} = [trapezium, trapezium angle=120, draw, rotate=90,  inner ysep=5pt, outer sep=5pt,
minimum height=7mm, minimum width=7mm]
\tikzstyle{roundbox} = [draw, circle, inner sep=0pt, minimum size=3mm]
\tikzstyle{clamped} = [draw, fill=darkgrey, minimum size=0.15cm]
\tikzstyle{msgcircle} = [shape=circle, draw, inner sep=0pt, minimum size=4mm, fill=white, font=\scriptsize]
\tikzstyle{darkmsgcircle} = [shape=circle, draw, inner sep=0pt, minimum size=4mm, fill=darkgrey, text=white, font=\scriptsize]
\tikzstyle{redmsgcircle} = [shape=circle, draw=red, inner sep=0pt, minimum size=4mm, text=red, font=\scriptsize]
\tikzstyle{reddarkmsgcircle} = [shape=circle, draw=red, inner sep=0pt, minimum size=4mm, fill=red, text=white, font=\scriptsize]
\tikzstyle{msgdoublecircle} = [shape=circle, double, double distance=1.5pt, draw, inner sep=0pt, minimum size=5mm, fill=white]
\tikzstyle{darkmsgdoublecircle} = [shape=circle, double, double distance=1.5pt, draw, inner sep=0pt, minimum size=5mm, fill=darkgrey, text=white, font=\bfseries]
\newcommand{\msg}[6]{
      \ifthenelse{\isin{#1}{left} \AND \isin{#2}{down}}{
            \coordinate (anchor) at ($({#3})!{#5}!({#4})$);
            \node[xshift=-6.0mm] at (anchor) {#6};
            \node[xshift=-1.0mm] at (anchor) {$\downarrow$};
      }{}
      \ifthenelse{\isin{#1}{right} \AND \isin{#2}{down}}{
            \coordinate (anchor) at ($({#3})!{#5}!({#4})$);
            \node[xshift=6.0mm] at (anchor) {#6};
            \node[xshift=1.0mm] at (anchor) {$\downarrow$};
      }{}

      \ifthenelse{\isin{#1}{down} \AND \isin{#2}{right}}{
            \coordinate (anchor) at ($({#3})!{#5}!({#4})$);
            \node[ yshift=-4.0mm] at (anchor) {#6};
            \node[yshift=-1.0mm] at (anchor) {$\rightarrow$};
      }{}
      \ifthenelse{\isin{#1}{up} \AND \isin{#2}{right}}{
            \coordinate (anchor) at ($({#3})!{#5}!({#4})$);
            \node[ yshift=4.0mm] at (anchor) {#6};
            \node[yshift=1.0mm] at (anchor) {$\rightarrow$};
      }{}

      \ifthenelse{\isin{#1}{down} \AND \isin{#2}{left}}{
            \coordinate (anchor) at ($({#3})!{#5}!({#4})$);
            \node[ yshift=-4.0mm] at (anchor) {#6};
            \node[yshift=-1.0mm] at (anchor) {$\leftarrow$};
      }{}
      \ifthenelse{\isin{#1}{up} \AND \isin{#2}{left}}{
            \coordinate (anchor) at ($({#3})!{#5}!({#4})$);
            \node[ yshift=4.0mm] at (anchor) {#6};
            \node[yshift=1.0mm] at (anchor) {$\leftarrow$};
      }{}

      \ifthenelse{\isin{#1}{left} \AND \isin{#2}{up}}{
            \coordinate (anchor) at ($({#3})!{#5}!({#4})$);
            \node[ xshift=-6.0mm] at (anchor) {#6};
            \node[xshift=-1.0mm] at (anchor) {$\uparrow$};
      }{}
      \ifthenelse{\isin{#1}{right} \AND \isin{#2}{up}}{
            \coordinate (anchor) at ($({#3})!{#5}!({#4})$);
            \node[ xshift=6.0mm] at (anchor) {#6};
            \node[xshift=1.0mm] at (anchor) {$\uparrow$};
      }{}
}

\newcommand{\msgcircle}[6]{
      \ifthenelse{\isin{#1}{left} \AND \isin{#2}{down}}{
            \coordinate (anchor) at ($({#3})!{#5}!({#4})$);
            \node[msgcircle,xshift=-5.0mm] at (anchor) {#6};
            \node[xshift=-1.5mm] at (anchor) {$\downarrow$};
      }{}
      \ifthenelse{\isin{#1}{right} \AND \isin{#2}{down}}{
            \coordinate (anchor) at ($({#3})!{#5}!({#4})$);
            \node[msgcircle,xshift=5.0mm] at (anchor) {#6};
            \node[xshift=1.5mm] at (anchor) {$\downarrow$};
      }{}

      \ifthenelse{\isin{#1}{down} \AND \isin{#2}{right}}{
            \coordinate (anchor) at ($({#3})!{#5}!({#4})$);
            \node[msgcircle, yshift=-5.0mm] at (anchor) {#6};
            \node[yshift=-2.0mm] at (anchor) {$\rightarrow$};
      }{}
      \ifthenelse{\isin{#1}{up} \AND \isin{#2}{right}}{
            \coordinate (anchor) at ($({#3})!{#5}!({#4})$);
            \node[msgcircle, yshift=5.0mm] at (anchor) {#6};
            \node[yshift=2.0mm] at (anchor) {$\rightarrow$};
      }{}

      \ifthenelse{\isin{#1}{down} \AND \isin{#2}{left}}{
            \coordinate (anchor) at ($({#3})!{#5}!({#4})$);
            \node[msgcircle, yshift=-5.0mm] at (anchor) {#6};
            \node[yshift=-2.0mm] at (anchor) {$\leftarrow$};
      }{}
      \ifthenelse{\isin{#1}{up} \AND \isin{#2}{left}}{
            \coordinate (anchor) at ($({#3})!{#5}!({#4})$);
            \node[msgcircle, yshift=5.0mm] at (anchor) {#6};
            \node[yshift=2.0mm] at (anchor) {$\leftarrow$};
      }{}

      \ifthenelse{\isin{#1}{left} \AND \isin{#2}{up}}{
            \coordinate (anchor) at ($({#3})!{#5}!({#4})$);
            \node[msgcircle, xshift=-5.0mm] at (anchor) {#6};
            \node[xshift=-1.5mm] at (anchor) {$\uparrow$};
      }{}
      \ifthenelse{\isin{#1}{right} \AND \isin{#2}{up}}{
            \coordinate (anchor) at ($({#3})!{#5}!({#4})$);
            \node[msgcircle, xshift=5.0mm] at (anchor) {#6};
            \node[xshift=1.5mm] at (anchor) {$\uparrow$};
      }{}
}

\newcommand{\darkmsg}[6]{
      \ifthenelse{\isin{#1}{left} \AND \isin{#2}{down}}{
            \coordinate (anchor) at ($({#3})!{#5}!({#4})$);
            \node[darkmsgcircle, xshift=-5mm] at (anchor) {#6};
            \node[xshift=-1.5mm] at (anchor) {$\downarrow$};
      }{}
      \ifthenelse{\isin{#1}{right} \AND \isin{#2}{down}}{
            \coordinate (anchor) at ($({#3})!{#5}!({#4})$);
            \node[darkmsgcircle, xshift=5mm] at (anchor) {#6};
            \node[xshift=1.5mm] at (anchor) {$\downarrow$};
      }{}

      \ifthenelse{\isin{#1}{down} \AND \isin{#2}{right}}{
            \coordinate (anchor) at ($({#3})!{#5}!({#4})$);
            \node[darkmsgcircle, yshift=-5.0mm] at (anchor) {#6};
            \node[yshift=-2.0mm] at (anchor) {$\rightarrow$};
      }{}
      \ifthenelse{\isin{#1}{up} \AND \isin{#2}{right}}{
            \coordinate (anchor) at ($({#3})!{#5}!({#4})$);
            \node[darkmsgcircle, yshift=5.0mm] at (anchor) {#6};
            \node[yshift=2.0mm] at (anchor) {$\rightarrow$};
      }{}

      \ifthenelse{\isin{#1}{down} \AND \isin{#2}{left}}{
            \coordinate (anchor) at ($({#3})!{#5}!({#4})$);
            \node[darkmsgcircle, yshift=-5.0mm] at (anchor) {#6};
            \node[yshift=-2.0mm] at (anchor) {$\leftarrow$};
      }{}
      \ifthenelse{\isin{#1}{up} \AND \isin{#2}{left}}{
            \coordinate (anchor) at ($({#3})!{#5}!({#4})$);
            \node[darkmsgcircle, yshift=5.0mm] at (anchor) {#6};
            \node[yshift=2.0mm] at (anchor) {$\leftarrow$};
      }{}

      \ifthenelse{\isin{#1}{left} \AND \isin{#2}{up}}{
            \coordinate (anchor) at ($({#3})!{#5}!({#4})$);
            \node[darkmsgcircle, xshift=-5.0mm] at (anchor) {#6};
            \node[xshift=-1.5mm] at (anchor) {$\uparrow$};
      }{}
      \ifthenelse{\isin{#1}{right} \AND \isin{#2}{up}}{
            \coordinate (anchor) at ($({#3})!{#5}!({#4})$);
            \node[darkmsgcircle, xshift=5.0mm] at (anchor) {#6};
            \node[xshift=1.5mm] at (anchor) {$\uparrow$};
      }{}
}

\newcommand{\redbackmsg}[6]{
      \ifthenelse{\isin{#1}{left} \AND \isin{#2}{down}}{
            \coordinate (anchor) at ($({#3})!{#5}!({#4})$);
            \node[reddarkmsgcircle, xshift=-5mm] at (anchor) {#6};
            \node[xshift=-1.5mm] at (anchor) {$\downarrow$};
      }{}
      \ifthenelse{\isin{#1}{right} \AND \isin{#2}{down}}{
            \coordinate (anchor) at ($({#3})!{#5}!({#4})$);
            \node[reddarkmsgcircle, xshift=5mm] at (anchor) {#6};
            \node[xshift=1.5mm] at (anchor) {$\downarrow$};
      }{}

      \ifthenelse{\isin{#1}{down} \AND \isin{#2}{right}}{
            \coordinate (anchor) at ($({#3})!{#5}!({#4})$);
            \node[reddarkmsgcircle, yshift=-5.0mm] at (anchor) {#6};
            \node[yshift=-2.0mm] at (anchor) {$\rightarrow$};
      }{}
      \ifthenelse{\isin{#1}{up} \AND \isin{#2}{right}}{
            \coordinate (anchor) at ($({#3})!{#5}!({#4})$);
            \node[reddarkmsgcircle, yshift=5.0mm] at (anchor) {#6};
            \node[yshift=2.0mm] at (anchor) {$\rightarrow$};
      }{}

      \ifthenelse{\isin{#1}{down} \AND \isin{#2}{left}}{
            \coordinate (anchor) at ($({#3})!{#5}!({#4})$);
            \node[reddarkmsgcircle, yshift=-5.0mm] at (anchor) {#6};
            \node[yshift=-2.0mm] at (anchor) {$\leftarrow$};
      }{}
      \ifthenelse{\isin{#1}{up} \AND \isin{#2}{left}}{
            \coordinate (anchor) at ($({#3})!{#5}!({#4})$);
            \node[reddarkmsgcircle, yshift=5.0mm] at (anchor) {#6};
            \node[yshift=2.0mm] at (anchor) {$\leftarrow$};
      }{}

      \ifthenelse{\isin{#1}{left} \AND \isin{#2}{up}}{
            \coordinate (anchor) at ($({#3})!{#5}!({#4})$);
            \node[reddarkmsgcircle, xshift=-5.0mm] at (anchor) {#6};
            \node[xshift=-1.5mm] at (anchor) {$\uparrow$};
      }{}
      \ifthenelse{\isin{#1}{right} \AND \isin{#2}{up}}{
            \coordinate (anchor) at ($({#3})!{#5}!({#4})$);
            \node[reddarkmsgcircle, xshift=5.0mm] at (anchor) {#6};
            \node[xshift=1.5mm] at (anchor) {$\uparrow$};
      }{}
}

\newcommand{\redmsg}[6]{
      \ifthenelse{\isin{#1}{left} \AND \isin{#2}{down}}{
            \coordinate (anchor) at ($({#3})!{#5}!({#4})$);
            \node[redmsgcircle, xshift=-5mm] at (anchor) {#6};
            \node[xshift=-1.5mm] at (anchor) {$\downarrow$};
      }{}
      \ifthenelse{\isin{#1}{right} \AND \isin{#2}{down}}{
            \coordinate (anchor) at ($({#3})!{#5}!({#4})$);
            \node[redmsgcircle, xshift=5mm] at (anchor) {#6};
            \node[xshift=1.5mm] at (anchor) {$\downarrow$};
      }{}

      \ifthenelse{\isin{#1}{down} \AND \isin{#2}{right}}{
            \coordinate (anchor) at ($({#3})!{#5}!({#4})$);
            \node[redmsgcircle, yshift=-5.0mm] at (anchor) {#6};
            \node[yshift=-2.0mm] at (anchor) {$\rightarrow$};
      }{}
      \ifthenelse{\isin{#1}{up} \AND \isin{#2}{right}}{
            \coordinate (anchor) at ($({#3})!{#5}!({#4})$);
            \node[redmsgcircle, yshift=5.0mm] at (anchor) {#6};
            \node[yshift=2.0mm] at (anchor) {$\rightarrow$};
      }{}

      \ifthenelse{\isin{#1}{down} \AND \isin{#2}{left}}{
            \coordinate (anchor) at ($({#3})!{#5}!({#4})$);
            \node[redmsgcircle, yshift=-5.0mm] at (anchor) {#6};
            \node[yshift=-2.0mm] at (anchor) {$\leftarrow$};
      }{}
      \ifthenelse{\isin{#1}{up} \AND \isin{#2}{left}}{
            \coordinate (anchor) at ($({#3})!{#5}!({#4})$);
            \node[redmsgcircle, yshift=5.0mm] at (anchor) {#6};
            \node[yshift=2.0mm] at (anchor) {$\leftarrow$};
      }{}

      \ifthenelse{\isin{#1}{left} \AND \isin{#2}{up}}{
            \coordinate (anchor) at ($({#3})!{#5}!({#4})$);
            \node[redmsgcircle, xshift=-5.0mm] at (anchor) {#6};
            \node[xshift=-1.5mm] at (anchor) {$\uparrow$};
      }{}
      \ifthenelse{\isin{#1}{right} \AND \isin{#2}{up}}{
            \coordinate (anchor) at ($({#3})!{#5}!({#4})$);
            \node[redmsgcircle, xshift=5.0mm] at (anchor) {#6};
            \node[xshift=1.5mm] at (anchor) {$\uparrow$};
      }{}
}

\newcommand{\bwmsg}[6]{
      \ifthenelse{\isin{#1}{left} \AND \isin{#2}{down}}{
            \coordinate (anchor) at ($({#3})!{#5}!({#4})$);
            \node[msgdoublecircle, xshift=-5.5mm] at (anchor) {#6};
            \node[xshift=-1.5mm] at (anchor) {$\downarrow$};
      }{}
      \ifthenelse{\isin{#1}{right} \AND \isin{#2}{down}}{
            \coordinate (anchor) at ($({#3})!{#5}!({#4})$);
            \node[msgdoublecircle, xshift=5.5mm] at (anchor) {#6};
            \node[xshift=1.5mm] at (anchor) {$\downarrow$};
      }{}

      \ifthenelse{\isin{#1}{down} \AND \isin{#2}{right}}{
            \coordinate (anchor) at ($({#3})!{#5}!({#4})$);
            \node[msgdoublecircle, yshift=-6.0mm] at (anchor) {#6};
            \node[yshift=-2.0mm] at (anchor) {$\rightarrow$};
      }{}
      \ifthenelse{\isin{#1}{up} \AND \isin{#2}{right}}{
            \coordinate (anchor) at ($({#3})!{#5}!({#4})$);
            \node[msgdoublecircle, yshift=6.0mm] at (anchor) {#6};
            \node[yshift=2.0mm] at (anchor) {$\rightarrow$};
      }{}

      \ifthenelse{\isin{#1}{down} \AND \isin{#2}{left}}{
            \coordinate (anchor) at ($({#3})!{#5}!({#4})$);
            \node[msgdoublecircle, yshift=-6.0mm] at (anchor) {#6};
            \node[yshift=-2.0mm] at (anchor) {$\leftarrow$};
      }{}
      \ifthenelse{\isin{#1}{up} \AND \isin{#2}{left}}{
            \coordinate (anchor) at ($({#3})!{#5}!({#4})$);
            \node[msgdoublecircle, yshift=6.0mm] at (anchor) {#6};
            \node[yshift=2.0mm] at (anchor) {$\leftarrow$};
      }{}

      \ifthenelse{\isin{#1}{left} \AND \isin{#2}{up}}{
            \coordinate (anchor) at ($({#3})!{#5}!({#4})$);
            \node[msgdoublecircle, xshift=-5.5mm] at (anchor) {#6};
            \node[xshift=-1.5mm] at (anchor) {$\uparrow$};
      }{}
      \ifthenelse{\isin{#1}{right} \AND \isin{#2}{up}}{
            \coordinate (anchor) at ($({#3})!{#5}!({#4})$);
            \node[msgdoublecircle, xshift=5.5mm] at (anchor) {#6};
            \node[xshift=1.5mm] at (anchor) {$\uparrow$};
      }{}
}

\newcommand{\bwdarkmsg}[6]{
      \ifthenelse{\isin{#1}{left} \AND \isin{#2}{down}}{
            \coordinate (anchor) at ($({#3})!{#5}!({#4})$);
            \node[darkmsgdoublecircle, xshift=-5.5mm] at (anchor) {#6};
            \node[xshift=-1.5mm] at (anchor) {$\downarrow$};
      }{}
      \ifthenelse{\isin{#1}{right} \AND \isin{#2}{down}}{
            \coordinate (anchor) at ($({#3})!{#5}!({#4})$);
            \node[darkmsgdoublecircle, xshift=5.5mm] at (anchor) {#6};
            \node[xshift=1.5mm] at (anchor) {$\downarrow$};
      }{}

      \ifthenelse{\isin{#1}{down} \AND \isin{#2}{right}}{
            \coordinate (anchor) at ($({#3})!{#5}!({#4})$);
            \node[darkmsgdoublecircle, yshift=-6.0mm] at (anchor) {#6};
            \node[yshift=-2.0mm] at (anchor) {$\rightarrow$};
      }{}
      \ifthenelse{\isin{#1}{up} \AND \isin{#2}{right}}{
            \coordinate (anchor) at ($({#3})!{#5}!({#4})$);
            \node[darkmsgdoublecircle, yshift=6.0mm] at (anchor) {#6};
            \node[yshift=2.0mm] at (anchor) {$\rightarrow$};
      }{}

      \ifthenelse{\isin{#1}{down} \AND \isin{#2}{left}}{
            \coordinate (anchor) at ($({#3})!{#5}!({#4})$);
            \node[darkmsgdoublecircle, yshift=-6.0mm] at (anchor) {#6};
            \node[yshift=-2.0mm] at (anchor) {$\leftarrow$};
      }{}
      \ifthenelse{\isin{#1}{up} \AND \isin{#2}{left}}{
            \coordinate (anchor) at ($({#3})!{#5}!({#4})$);
            \node[darkmsgdoublecircle, yshift=6.0mm] at (anchor) {#6};
            \node[yshift=2.0mm] at (anchor) {$\leftarrow$};
      }{}

      \ifthenelse{\isin{#1}{left} \AND \isin{#2}{up}}{
            \coordinate (anchor) at ($({#3})!{#5}!({#4})$);
            \node[darkmsgdoublecircle, xshift=-5.5mm] at (anchor) {#6};
            \node[xshift=-1.5mm] at (anchor) {$\uparrow$};
      }{}
      \ifthenelse{\isin{#1}{right} \AND \isin{#2}{up}}{
            \coordinate (anchor) at ($({#3})!{#5}!({#4})$);
            \node[darkmsgdoublecircle, xshift=5.5mm] at (anchor) {#6};
            \node[xshift=1.5mm] at (anchor) {$\uparrow$};
      }{}
}

\tikzset{mainstyle/.style={fill=white, draw=black, shape=rectangle, align=center}}

\tikzset{dstyle/.style={mainstyle, minimum size=4mm, inner sep=0pt, text width=4mm}}

\tikzset{sstyle/.style={mainstyle, minimum size=5mm, inner sep=0pt, text width=5mm}}

\tikzset{ostyle/.style={fill=darkgrey, draw=black, shape=rectangle, minimum size=0.2cm, inner sep=0pt, text width=2mm}}

\tikzstyle{observation}=[ostyle]
\tikzstyle{deterministic}=[dstyle]
\tikzstyle{stochastic}=[sstyle]

\tikzstyle{filter}=[mainstyle, minimum width=1cm, minimum height=0.5cm]
\tikzstyle{selector}=[fill=white, draw=black, shape=trapezium, rotate=180, minimum width=1cm, minimum height=0.5cm]

\DeclareRobustCommand{\cev}[1]{%
  \mathpalette\do@cev{#1}%
}
\newcommand{\do@cev}[2]{%
  \fix@cev{#1}{+}%
  \reflectbox{$\m@th#1\vec{\reflectbox{$\fix@cev{#1}{-}\m@th#1#2\fix@cev{#1}{+}$}}$}%
  \fix@cev{#1}{-}%
}
\newcommand{\fix@cev}[2]{%
  \ifx#1\displaystyle
    \mkern#23mu
  \else
    \ifx#1\textstyle
      \mkern#23mu
    \else
      \ifx#1\scriptstyle
        \mkern#22mu
      \else
        \mkern#22mu
      \fi
    \fi
  \fi
}
\newcommand\given[1][]{\:#1\vert\:}

\usepackage[]{todonotes}

\theoremstyle{thmstyleone}%
\theoremstyle{thmstyletwo}%

\theoremstyle{thmstylethree}%

\begin{document}

% \title[STDP-based Spiking Neural Networks
% for Belief Propagation]{STDP-based Spiking Neural Networks
% for Belief Propagation}

\title[Spiking Neural Network Implementation of Gaussian Belief Propagation]{A Spiking Neural Network Implementation of Gaussian Belief Propagation}


\author*{\fnm{Sepideh} \sur{Adamiat}}\email{s.adamiat@tue.nl}

\author{\fnm{Wouter M.} \sur{Kouw}}\email{w.m.kouw@tue.nl}

\author{\fnm{Bert} \spfx{de}  \sur{Vries}}\email{bert.de.vries@tue.nl}

%\affil*[1]{\orgdiv{Electrical Engineering Department}, \orgname{Eindhoven University of Technology}, \orgaddress{\street{De Groene Loper}, \city{Eindhoven}, \postcode{5612 AP}, \state{North Brabant}, \country{Netherlands}}}

\affil{\orgdiv{Electrical Engineering Department}, \orgname{Eindhoven University of Technology}, \city{Eindhoven}, \country{the Netherlands}}

\abstract{
Bayesian inference offers a principled account of information processing in natural agents. However, it remains an open question how neural mechanisms perform their abstract operations. 
We investigate a hypothesis where a distributed form of Bayesian inference, namely message passing on factor graphs, is performed by a simulated network of leaky-integrate-and-fire neurons. Specifically, 
we perform Gaussian belief propagation by encoding messages that come into factor nodes as spike-based signals, propagating these signals through a spiking neural network (SNN) and decoding the spike-based signal back to an outgoing message. 
Three core linear operations, equality (branching), addition, and multiplication, are realized in networks of leaky integrate-and-fire models. 
Validation against the standard sum-product algorithm shows accurate message updates, while applications to Kalman filtering and Bayesian linear regression demonstrate the framework’s potential for both static and dynamic inference tasks. Our results provide a step toward biologically grounded, neuromorphic implementations of probabilistic reasoning.}


\maketitle

% \listoftodos

% \newpage\tableofcontents \newpage

\section{Introduction}\label{sec:Introduction}

\begin{figure}[H]
    \centering
    \includegraphics[width=0.95\textwidth]{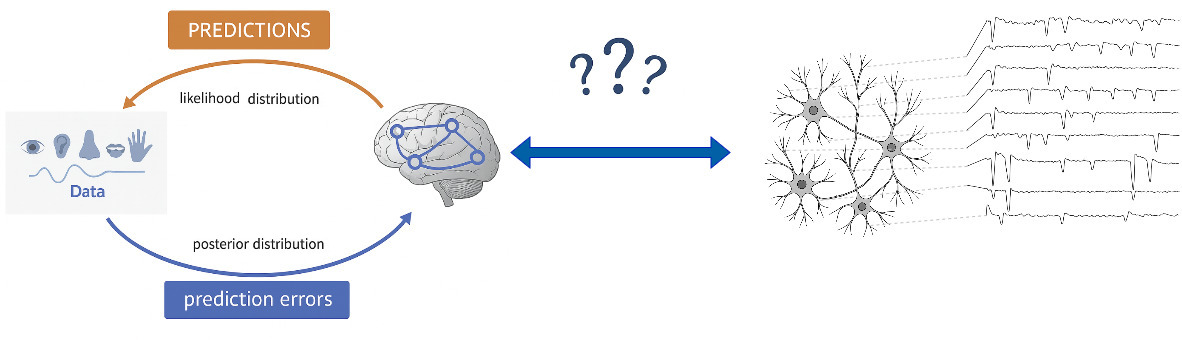}
    \caption{The Bayesian Brain Hypothesis (left) suggests that the brain performs probabilistic inference by predicting environmental states and updating these predictions using sensory input. On the right, neural spike activity depicts the brain’s underlying biological dynamics. This paper proposes a method to implement Bayesian inference through spiking activity, thereby linking abstract probabilistic formulations to biologically plausible neural dynamics.}
    \label{fig:idea}
\end{figure}

In recent years, brain-inspired algorithms have attracted increasing attention for their potential to advance both artificial intelligence and neuroscience. A central theory of brain function posits that the brain performs computations consistent with Bayes’ theorem, allowing it to represent and manipulate uncertainty in a principled manner~\cite{knill2004bayesian, friston2005theory}. 
The Bayesian framework has been used to describe the information processing of the central nervous system for many perceptual and motor tasks \cite{kording2004bayesian,knill1996perception}. Despite the appeal of this framework, a substantial gap remains between the abstract formulations of Bayesian computation and the biological mechanisms underlying neural activity (see Figure~\ref {fig:idea}).
In this paper, we aim to narrow this gap by presenting a novel neuro-inspired Bayesian inference implementation.

Our work is based on the use of spiking neural networks (SNNs)~\cite{maass1997networks}. Unlike traditional artificial neural networks, SNNs process information through exchanging spike trains between neurons. SNNs exhibit rich spatiotemporal dynamics, leverage diverse neural coding schemes, and operate in an inherently event-driven manner, which makes them highly energy-efficient when implemented on dedicated neuromorphic hardware such as Intel’s Loihi chip~\cite{davies2018loihi}.
Several studies have proposed SNN architectures to implement Bayesian inference~\cite{pecevski2011probabilistic, deneve2004bayesian, vafaii2024brain, litvak2009cortical}. In contrast to previous work, we focus on a distributed form of Bayesian inference, namely message passing on factor graphs ~\cite{kschischang2002factor}. Unlike brute-force approaches, where the cost of computing marginal distributions grows exponentially with the number of variables, message passing algorithms like Belief Propagation (BP) decompose a global inference task into local operations between neighboring nodes. This distributed structure enables inference to scale to large models while also reflecting the parallel and decentralized organization of biological neural processing, where individual units communicate only through local connections.
In \cite{ott2006neurodynamics}, a close relationship between BP and neuro-dynamic models was established. However, their work was limited to Markov random fields, which represent only a subset of factor graphs and are restricted to binary variables. An extension was presented by \cite{steimer2009belief}, who implemented BP in SNNs using a network of interconnected liquid state machines~\cite{maass2011liquid}. Their results provide valuable insights into how BP can be realized in biologically inspired systems for broader classes of problems. However, important aspects remain unexplored, most notably extending the message update mechanism beyond Bernoulli distributions and providing synaptic learning processes. To the best of our knowledge, no existing model supports spike-encoded Gaussian message updates for BP, which is the goal of this paper.

Our overall aim is to bridge two successful research directions in neural implementations of Bayesian inference. Within the signal processing and information theory communities, factor graphs are a well-established framework to efficiently implement Bayesian inference algorithms \cite{loeliger_factor_2007}. In the neuro-inspired AI community, SNNs have been developed as biologically plausible computational structures for information processing. By leveraging SNNs for spike-encoded Gaussian belief updating, our approach advances the feasibility of applying bio-inspired belief propagation to complex inference problems.

In Section~\ref{sec:background}, we review the technical background related to message passing in factor graphs, as well as relevant materials for the construction of SNNs. In particular, we review the Leaky-Integrate-and-Fire (LIF) neuron model, as well as the spike-timing-dependent plasticity (STDP) protocol, which is a biologically inspired synaptic update rule that is widely used in neuromorphic computing. 
Section~\ref{section:solution} presents our solution proposal: an implementation of spike-encoded Gaussian message passing in SNN models. In particular, we derive message update rules for equality (branching), summation, and multiplication nodes in a factor graph, thereby providing a complete framework for message passing–based Bayesian inference. 
We show proof-of-concept validation experiments for Bayesian linear regression and Kalman filtering tasks in Section~\ref{section:validation}.

\section{Problem Statement}\label{sec:problem-statement}

Bayesian inference and neural activity describe two fundamentally different levels of explanation for information processing in the brain. Bayesian inference is a principled way to update what we believe about something, such as a hypothesis, a hidden cause, or an unobserved variable, after receiving new evidence. It provides a formal mathematical framework for reasoning under uncertainty by combining prior knowledge with observed data. Consider two random variables, $x$ and $y$, where $y$ is observed and $x$ is latent. The observation $y$ carries information about the latent variable $x$. Bayes’ rule prescribes how to update our belief about $x$ from a prior to a posterior distribution~\cite{bishop2006pattern}
\begin{align}
    \underbrace{p(x \given y)}_{\text{posterior}} = \frac{\overbrace{p(y \given x)}^{\text{likelihood}}}{\underbrace{p(y)}_{\text{evidence}}} \underbrace{p(x)}_{\text{prior}} \, .
\end{align}
In essence, Bayesian inference formalizes how rational agents, or biological systems, should revise their beliefs when confronted with new observations, allowing them to make probabilistic judgments that incorporate both prior expectations and sensory evidence.

In contrast, biological neural systems do not explicitly represent or manipulate probability distributions. Instead, information is encoded and processed through dynamic patterns of neural activity, such as the firing rates of individual neurons $r_i$, membrane potentials $v_i$, and between neurons $i$ and $j$, denoted $w_{ij}$. The central challenge, therefore, is to understand how these biophysical variables in a population of $N$ neurons can encode the brain’s internal beliefs over latent variables $x$, and how this representation is updated by sensory input $y$. 

Whereas the brain performs inference continuously through sparse, robust, and energy-efficient dynamics, conventional implementations of Bayesian inference on CPUs lack these key properties. Bridging the gap between these frameworks could yield mutual benefits: enabling more biologically inspired, low-power Bayesian algorithms and deepening our understanding of neural computation.  

We present the technical background in the next section. Readers familiar with this material may proceed directly to Section~\ref{section:solution}.

\section{Technical Background}\label{sec:background}

In this section, we review two powerful frameworks for neuro-inspired artificial intelligence systems: Bayesian inference through message passing and SNNs. 

\subsection{Factor Graphs and Message Passing-based Inference}\label{sec:FFG-review}

To implement Bayesian inference in a computationally efficient and scalable manner, we employ a factor graph representation. A factor graph provides a graphical representation of a factorized probabilistic model, allowing Bayesian inference to be performed through the propagation of beliefs(or message passing) via local computations. This formulation enhances scalability and modularity while offering a natural fit for neural computation, where local interactions among neurons may play roles analogous to message exchanges in a factor graph.

In this work, we focus on Forney-style Factor Graphs (FFGs), where edges correspond to variables and nodes represent factors describing probabilistic relationships between the connected variables. In an FFG, a variable is connected to a node if and only if it appears as an argument of the node’s function~\cite{forney2001codes}. Since an edge can connect to at most two nodes, the FFG framework introduces branching nodes, also known as equality nodes, which enable the representation of models in which a variable participates in three or more factors~\cite{loeliger2004introduction}.

We proceed by an example. Consider the task of tracking the position of a car over time, where both the evolution and the measurements are corrupted by random noise. Let $u_t$ denote an external force, $x_t$ be the (unobserved) position of the car, and $y_t$ be a noisy observation of the car's position. We assume the following probabilistic generative model for time step $t$, 
\begin{align}
p(y_t,x_t,u_t \mid  x_{t-1}) &= p(y_t \mid x_t) p(x_t\mid x_{t-1}, u_t) p(u_t) \notag \\
&= \underbrace{\mathcal{N}(y_t\mid x_t,\sigma_y^2)}_{\text{observation}} \underbrace{\mathcal{N}(x_t\mid x_{t-1} + u_t,\sigma_x^2)}_{\text{process dynamics}} \underbrace{\mathcal{N}(u_t\mid m_u,\sigma_u^2)}_{\text{driving force}} \,. \label{eq:KF_model1}
\end{align}

Note that the joint distribution over all variables over a time interval $[0,T]$ is simply the product of the distributions defined at each individual time step,
\begin{align}
p(y_{1:T},x_{0:T},u_{1:T}) = p(x_0) \prod_{t=1}^T
p(y_t,x_t,u_t\mid x_{t-1}) \,.\label{eq:KF_model}
\end{align}

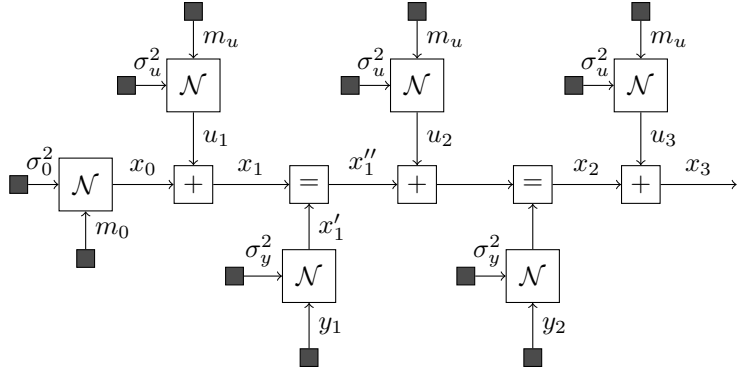
\begin{figure}[bth]
\centering
\begin{tikzpicture}
    %time step 1
    \node[smallbox] (sum1) {$+$};
    \node[smallbox, right = 10 mm of sum1] (eq1) {$=$};
    \node[box, above = 7 mm of sum1] (v1) {$\mathcal{N}$};
    \node[clamped, above = 5 mm of v1] (m_v1) {};
    \node[clamped, left = 4 mm of v1] (v_v1) {};
    \node[box, below = 6 mm of eq1] (N1) {$\mathcal{N}$};
    \node[clamped, left = 5 mm of N1] (e1) {};
    \node[clamped, below = 6 mm of N1] (y1) {};
    \node[box,left = 8 mm of sum1] (left) {$\mathcal{N}$};
    \node[clamped, below = 5 mm of left] (m_s0) {};
    \node[clamped, left = 4 mm of left] (v_s0) {};
    
    \draw[->] (sum1) -- (eq1)node[pos=0.5, above] {$x_1$};
    \draw[<-] (sum1) -- (v1)node[pos=0.5, right] {$u_1$};
    \draw[->] (m_v1) -- (v1)node[pos=0.5, right] {$m_u$};
    \draw[->] (v_v1) -- (v1)node[pos=0.4, above] {$\sigma^2_u$};
    \draw[<-] (sum1) -- (left)node[pos=0.5, above] {$x_0$};
    \draw[->] (N1) -- (eq1)node[pos=0.5, right] {$x'_1$};
    \draw[<-] (N1) -- (e1)node[pos=0.6, above] {$\sigma^2_{y}$};
    \draw[<-] (N1) -- (y1)node[pos=0.5, right] {$y_1$};
    \draw[->] (v_s0) -- (left)node[pos=0.4, above] {$\sigma^2_{0}$};
    \draw[->] (m_s0) -- (left)node[pos=0.5, right] {$m_{0}$};

    % %time step 2
    \node[smallbox, right = 9 mm of eq1] (sum2) {$+$};
    \node[smallbox, right = 10 mm of sum2] (eq2) {$=$};
    \node[box, above  = 7 mm of sum2] (v2) {$\mathcal{N}$};
    \node[clamped, above = 5 mm of v2] (m_v2) {};
    \node[clamped, left = 4 mm of v2] (v_v2) {};
    \node[box, below = 6 mm of eq2] (N2) {$\mathcal{N}$};
    \node[clamped, left = 4 mm of N2] (e2) {};
    \node[clamped, below = 6 mm of N2] (y2) {};

    \draw[->] (eq1) -- (sum2)node[pos=0.5, above] {$x''_1$};
    \draw[->] (sum2) -- (eq2);
    \draw[<-] (sum2) -- (v2)node[pos=0.5, right] {$u_2$};
    \draw[->] (m_v2) -- (v2)node[pos=0.5, right] {$m_u$};
    \draw[->] (v_v2) -- (v2)node[pos=0.4, above] {$\sigma^2_u$};
    \draw[->] (N2) -- (eq2);
    \draw[<-] (N2) -- (e2)node[pos=0.6, above] {$\sigma^2_{y}$};
    \draw[<-] (N2) -- (y2)node[pos=0.5, right] {$y_{2}$};

    %time step T
    \node[smallbox, right = 9 mm of eq2] (sumT) {$+$};
    \node[box, above  = 7 mm of sumT] (vT) {$\mathcal{N}$};
    \node[clamped, above = 5 mm of vT] (m_vT) {};
    \node[clamped, left = 4 mm of vT] (v_vT) {};
    \node[right = 5 mm and 10 mm of sumT] (NT) {};
    
    \draw[<-] (sumT) -- (vT)node[pos=0.5, right] {$u_3$};
    \draw[->] (sumT) -- (NT)node[pos=0.5, above] {$x_3$};
    \draw[<-] (sumT) -- (eq2)node[pos=0.5, above] {$x_2$};
    \draw[->] (m_vT) -- (vT)node[pos=0.5, right] {$m_u$};
    \draw[->] (v_vT) -- (vT)node[pos=0.4, above] {$\sigma^2_u$};

\end{tikzpicture}
\caption{FFG for the car's position tracking over three time steps. At each step, the posterior distribution $p(x_t \mid y_{1:t})$ is updated from 
the previous estimate $p(x_{t-1} \mid y_{1:t-1})$ and the new observation $y_t$. 
}
\label{fig:FFG-KF}
\end{figure}

The FFG for the generative model \eqref{eq:KF_model} for $3$ time steps (with individual time step models from \eqref{eq:KF_model1}) is shown in Fig.~\ref{fig:FFG-KF}.

The graph comprises four types of nodes. A data node, indicated by a small black box, denotes a known value of the associated variable. In an FFG, the factor associated with a node is defined as a function of its connected variables. Hence, for an observed value $y_1 = 5$, the factor of this data node is given by $f(y_1) = \delta(y_1 - 5)$, where $\delta(\cdot)$ is a Dirac delta function.
The factor for the Gaussian node that connects to $w$, indicated by $\mathcal{N}$, is 
\begin{equation}
    f_\mathcal{N}(u_1,m_u,\sigma^2_u) = \mathcal{N}(u_1\mid m_u,\sigma^2_u) \,.
\end{equation}
The other Gaussian nodes follow the same idea. 
The equality node for $x_1$ is given by
\begin{equation}
    f_=(x_1,x_1',x_1'') = \delta(x_1-x_1') \delta(x_1-x_1'') \,.
\end{equation}
This equality node ensures that the posterior beliefs over $x_1$, $x_1'$, and $x_1''$ will be the same. 
Finally, Fig.~\ref{fig:FFG-KF} contains summation nodes, that implement the factors
\begin{equation}
    f_{+}(x_t,x_{t-1},u_t) = \delta(x_t - (x_{t-1} + u_t)) \,.
\end{equation}
 Generally, any linear system with observations and Gaussian uncertainties can be represented as an FFG by a connected network of data, Gaussian, equality, addition, and multiplication nodes. 

\begin{figure}[bt]
    \centering
    \begin{tikzpicture}[
    box/.style={draw, rectangle, minimum size=1.0cm, thick, align=center},
    node distance=0.8cm and 1.2cm
]
% Function Node (Processing Box)
\node[box] (g) {\large $f$};

% Inputs (x1 shifted slightly upward)
\node[left=1.4cm of g, yshift=0.5cm] (x1){};
% \node[below left = 0.1cm of x1] (dots){$$\vdots$$};
\node[below right=0.01cm and 0.1cm of x1] (dots) {$\vdots$};
\node[below=1cm of x1] (xn){};

% Output (y)
\node[right=1.2cm of g] (y){};

% Connections with Message Labels
\draw[->, thick] (x1) -- (g) node[pos=0.5, above] {$x_1$};
\draw[->, thick] (xn) -- (g) node[pos=0.5, above] {$x_n$}; 
\draw[->, thick] (g) -- (y) node[pos=0.5, above] {$y$};

\end{tikzpicture}

% \begin{tikzpicture}
% \node[box] (g) {$f$};
% \node[below = 5 mm  of g] (down) {};
% \node[left = 8 mm  of g] (left) {};
% \node[right = 8 mm  of g] (right) {};

% \draw[->] (left) -- (g)node[pos=0.5, above] {$x$};
% \draw[->] (g) -- (right)node[pos=0.5, above] {$z$};
% \draw[<-] (g) -- (down)node[pos=0.5, right] {$y$};

% \end{tikzpicture}
    \caption{Node example related to the sum-product update rule in \eqref{eq:sum-product}, specifying the relation of input messages $\overrightarrow{\mu}_{x_1}(x_1)...\overrightarrow{\mu}_{x_n}(x_n)$ and the output message $\overrightarrow{\mu}_{y}(y)$.}
    \label{fig:node}
\end{figure}
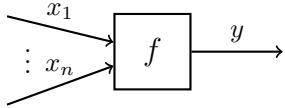

Inference in an FFG can be interpreted as a message passing process. If the graph is a tree, messages proceed from the terminals of the graph toward the target variable (in our example: $x_2$). 
Note that an edge variable may receive messages from both directions, depending on the inference task (as will be demonstrated later in the linear regression example). In order to distinguish these ``colliding" messages, we give the edges in the graph a direction. A message in the same direction as the edge direction is called a forward message, and a message in the opposite direction is called a backward message. 
It can be shown that Bayesian inference implies that, for a general node $f(y, x_1, \ldots, x_n)$ (see Fig.~\ref{fig:node}), with incoming messages $\overrightarrow{\mu}_{x_i}(x_i)$, the outgoing message toward $y$ is given by \cite{kschischang2002factor,loeliger2004introduction}
\begin{equation} \underbrace{\overrightarrow\mu_y({y})}_{\text{outgoing message}}=\int\underbrace{\overrightarrow\mu_{x_1}({x}_1)...\overrightarrow\mu_{x_n}({x}_n) }_{\text{incoming messages}} \underbrace{f({y},{x}_1,...,{x}_n)}_{\text{node function}} \mathrm{d}{x}_1...\mathrm{d}{x}_n \, . \label{eq:sum-product} 
\end{equation}
The message rule in \eqref{eq:sum-product} is referred to as the sum–product rule. Message passing based on sum–product rules is called Belief Propagation (BP), which yields exact Bayesian inference results when the underlying graph is a tree. Even when the graph contains cycles, iterative belief propagation often yields very good approximate inference results. 

Inference via message passing is computationally efficient as it only requires examining direct variable dependencies. From a technical perspective, message passing leverages the distributive law, transforming a computationally expensive sum-of-products into a typically much cheaper product-of-sums ~\cite{bishop2006pattern,aji2002generalized}. Message passing–based inference can be extended to variational inference and finds applications wherever factorized models are used, including (nonlinear) control problems, multi-agent trajectory planning, inference in microgrid systems, and signal processing~\cite{adamiat2024message,van2024multi,hu2010micro,colavolpe2005application}. Moreover, many well-known algorithms in machine learning, signal processing, coding theory, and statistics can be interpreted as special cases of message passing based on the sum–product rule~\cite{loeliger_factor_2007,palmieri2022unifying}.
Modern probabilistic programming packages, such as Infer.NET~\cite{InferNET18} and RxInfer.jl~\cite{bagaev2023reactive}, provide closed-form solutions for message update rules for many common node types. These packages enable the automation of Bayesian inference for a given model and data set with a single command. Update rules for forward and backward messages in Gaussian linear systems are shown in Table~\ref{table:mp-n}. 

As an example, we derive the outgoing message $\overrightarrow{\mu}_z(z)$ for the equality node $f(x,y,z) = \delta(z-x) \delta(z-y)$, when the incoming messages are Gaussian distributed, $\overrightarrow{\mu}_x(x) = \mathcal{N}(x|m_x,\sigma_x^2)$ and $\overrightarrow{\mu}_y(y) = \mathcal{N}(y|m_y,\sigma_y^2)$. The application of the sum-product rule \eqref{eq:sum-product} yields
\begin{subequations}
  \begin{align}
   \overrightarrow{\mu}_z(z) &=  \iint \overbrace{\overrightarrow{\mu}_x(x) \overrightarrow{\mu}_y(y)}^{\text{incoming messages}} \overbrace{f(x,y,z)}^{\text{node}} \mathrm{d}x  \mathrm{d}y \\
   &= \iint \mathcal{N}(x\mid m_x,\sigma_x^2) \mathcal{N}(y\mid m_y,\sigma_y^2)  \delta(z-x) \delta(z-y) \mathrm{d}x  \mathrm{d}y \\
   &= \mathcal{N}(z\mid m_x,\sigma_x^2) \mathcal{N}(z\mid m_y,\sigma_y^2) \\
   &\propto \mathcal{N}(z\,\mid \,m_z,\sigma_z^2) \,,
\end{align}   
\end{subequations}  
where 
\begin{align} \label{eq:eq-node-forward}
 \frac{1}{\sigma_z^2} = \frac{1}{\sigma_x^2} + \frac{1}{\sigma_y^2} \, , \qquad 
 \frac{1}{\sigma_z^2}m_z = \frac{1}{\sigma_x^2}m_x + \frac{1}{\sigma_y^2}m_y \, .
\end{align}      
Note that precisions (inverse variances) and precision-weighted means add when we multiply two Gaussian functions \cite{loeliger_factor_2007}. 
To execute automated inference, packages such as RxInfer.jl and Infer.NET store the result \eqref{eq:eq-node-forward} in a lookup table, which is accessed whenever Gaussian input messages arrive at an equality node. The other results in Table~\ref{table:mp-n} can be evaluated in a similar fashion.
For a more in-depth description of message passing-based inference, we recommend \cite{loeliger_factor_2007}.

\begin{table}[tbp]
    \centering
    \setlength{\tabcolsep}{6pt}
    \renewcommand{\arraystretch}{2.2}
    \caption{Sum-product update rules for linear operations on Gaussian messages, as described in \cite{loeliger_factor_2007}. Forward messages are computed given the input messages $\overrightarrow{\mu}_x(x) = \mathcal{N}(x \mid m_x, \sigma^2_x)$ and $\overrightarrow{\mu}_y(y) = \mathcal{N}(y \mid m_y, \sigma^2_y)$. The backward message on variable $y$ is computed by the inputs $\overrightarrow{\mu}_x(x) = \mathcal{N}(x \mid m_x, \sigma^2_x)$ and $\overleftarrow{\mu}_z(z) = \mathcal{N}(z \mid m_z, \sigma^2_z)$. Similarly, a backward message on $x$ can be derived for equality and summation nodes from symmetry arguments. In the multiplication node, $a$ is assumed to be a scalar and the backward message toward $a$ is not useful. 
    % \bdv{Add the node functions to the left boxes. The symbols do not fully describe the node function.} 
    }
    \vspace{5pt}
    \label{table:mp-n}
    \begin{tabular}{|c|c|c|}
        \hline
        \textbf{\textit{Node function}} & \textbf{\textit{Forward Update rule}} & \textbf{\textit{Backward Update rule}}\\
        \hline
        \begin{tabular}{c} \raisebox{-20pt}{\begin{tikzpicture}
\node[smallbox] (g) {$=$};
\node[below = 5 mm  of g] (down) {};
\node[left = 8 mm  of g] (left) {};
\node[right = 8 mm  of g] (right) {};

% \draw[-] (left) -- (g)node[pos=0.3, above] {$\overrightarrow{\mu}_x(x)$};
% \draw[-] (g) -- (right)node[pos=0.7, above] {$\overrightarrow{\mu}_z(z)$};
% \draw[-] (g) -- (down)node[pos=0.5, right] {$\uparrow\overrightarrow{\mu}_y(y)$};

\draw[->] (left) -- (g)node[pos=0.5, above] {$x$};
\draw[->] (g) -- (right)node[pos=0.5, above] {$z$};
\draw[<-] (g) -- (down)node[pos=0.5, right] {$y$};

\end{tikzpicture}} \\  $f_=(x,y,z) = \delta(z-x) \delta(z-y)$ \end{tabular}& 
        \begin{tabular}{c} 
            $\overrightarrow{\mu}_z(z) = \mathcal{N}(z \mid m_z , \sigma^2_z)$\\
            $ m_z = \sigma_z^2 \Big( \frac{m_x}{\sigma_x^2} + \frac{m_y}{\sigma_y^2} \Big)$ \\
            $\sigma_z^2 = \Big( \frac{1}{\sigma_x^2} + \frac{1}{\sigma_y^2} \Big)^{-1}$ \\ 
        \end{tabular} & 
        \begin{tabular}{c} 
            $\overleftarrow{\mu}_y(y) = \mathcal{N}(y \mid m_y , \sigma^2_y)$\\
            $m_y = \frac{w_x m_x + w_z m_z}{w_x + w_z}$ \\
            $\sigma^2 = w_x + w_z$ \\ 
        \end{tabular}\\ 
        \hline
        \begin{tabular}{c} \raisebox{-20pt}{\begin{tikzpicture}
\node[smallbox] (g) {$+$};
\node[below = 5 mm  of g] (down) {};
\node[left = 8 mm  of g] (left) {};
\node[right = 8 mm  of g] (right) {};

% \draw[->] (left) -- (g)node[pos=0.3, above] {$\overrightarrow{\mu}_x(x)$};
% \draw[->] (g) -- (right)node[pos=0.7, above] {$\overrightarrow{\mu}_z(z)$};
% \draw[<-] (g) -- (down)node[pos=0.5, right] {$\uparrow\overrightarrow{\mu}_y(y)$};

\draw[->] (left) -- (g)node[pos=0.5, above] {$x$};
\draw[->] (g) -- (right)node[pos=0.5, above] {$z$};
\draw[<-] (g) -- (down)node[pos=0.5, right] {$y$};

\end{tikzpicture}} \\ $f_+(x,y,z) = \delta(z-(x+y))$ \end{tabular} &  
        \begin{tabular}{c} 
            $\overrightarrow{\mu}_z(z) = \mathcal{N}(z \mid m_z , \sigma^2_z)$\\
            $m_z = m_x + m_y$ \\
            $\sigma^2_z = \sigma^2_x + \sigma^2_y$ \\ 
        \end{tabular} &
        \begin{tabular}{c} 
            $\overleftarrow{\mu}_y(y) = \mathcal{N}(y \mid m_y , \sigma^2_y)$\\
            $m_y = -m_x + m_z$ \\ 
            $\sigma^2_y = \sigma^2_x + \sigma^2_z$ \\ 
        \end{tabular} \\
        \hline
        \begin{tabular}{c} \raisebox{-20pt}{\begin{tikzpicture}
\node[smallbox] (g) {$\times$};
\node[below = 5 mm  of g] (down) {};
\node[left = 6 mm  of g] (left) {};
\node[right = 8 mm  of g] (right) {};
\node[clamped, left = 6 mm  of g] (a) {};

% \draw[->] (left) -- (g)node[pos=0.3, above] {$\overrightarrow{\mu}_x(x)$};
% \draw[->] (g) -- (right)node[pos=0.7, above] {$\overrightarrow{\mu}_z(z)$};
% \draw[<-] (g) -- (down)node[pos=0.5, right] {$\uparrow\overrightarrow{\mu}_y(y)$};

\draw[->] (left) -- (g)node[pos=0.5, above] {$a$};
\draw[->] (g) -- (right)node[pos=0.5, above] {$z$};
\draw[<-] (g) -- (down)node[pos=0.5, right] {$y$};

\end{tikzpicture}} \\ $f_{\times}(x,y,z) = \delta(z-ay)$\end{tabular} &  
        \begin{tabular}{c} 
            $\overrightarrow{\mu}_z(z) = \mathcal{N}(z  \mid  m_z , \sigma^2_z)$\\  
            $m_z = a m_y$ \\   
            $\sigma^2_z = a^2 \sigma^2_z$ \\ 
        \end{tabular} &
        \begin{tabular}{c} 
            $\overleftarrow{\mu}_y(y) = \mathcal{N}(y \mid m_y , \sigma^2_y)$\\   
            $m_x = \frac{1}{a}m_z$ \\
            $\sigma^2_x = \frac{1}{a^2}\sigma^2_z$ \\ 
        \end{tabular} \\
        \hline
    \end{tabular}
\end{table}

\subsection{Spiking Neural Networks}\label{sec:SNN}
SNNs represent a biologically inspired class of artificial neural networks that more closely emulate the dynamics of real neural systems than conventional ANNs. Unlike standard ANNs, which rely on continuous-valued activations and operate in a synchronous fashion, SNNs encode and transmit information through discrete spike events distributed over time. 
In this section, we describe the fundamental components of SNNs, with a focus on neuron models that produce spikes based on membrane potential dynamics and spike-encoded synaptic update models.

\subsubsection{The Leaky Integrate-and-Fire Neuron Model}\label{sec:LIF}

SNNs employ neuron models that describe the dynamics of membrane potentials and spike generation. A spike at time $t=\hat{t}$ is typically modeled as a Dirac delta function, 
\begin{align}
&\delta(t - \hat{t}) =
\begin{cases}
\infty  &\text{if } t = \hat{t} \\
0       &\text{if } t \ne \hat{t} 
\end{cases} \quad \text{, such that }\int_{-\infty}^{\infty} \delta(t - \hat{t}) \mathrm{d}t = 1 \, .
\end{align}

One of the most widely used neuron models is the Leaky Integrate-and-Fire (LIF) model, which simplifies the complex behavior of biological neurons while retaining the essential aspects of their behavior. The model is based on the membrane potential $V(t)$ at time $t$, whose dynamics can be described by
\begin{equation}\label{eq:lif-voltage}
    \tau_m\frac{dV(t)}{dt} = -\left(V(t) - V_r\right) + R_m I(t)\,,
\end{equation}
where $\tau_m$ is the membrane time constant, $V_r$ is a resting potential, $R_m$ is the membrane resistance, and $I(t)$  is the total synaptic input current \cite{gerstner2002spiking}. Current $I(t)$ can be modeled as the weighted sum of presynaptic spike arrivals convolved with synaptic response kernels,
\begin{equation}\label{eq:lif-current}
    I(t) = \sum_j w_j \epsilon(t - \hat{t}_j)
\end{equation}
 where $w_j$ are synaptic weights, $\hat{t}_j$ are the spiking times of input neurons $j$, and $\epsilon(\cdot)$ is the shape of the postsynaptic current, which can be interpreted as an impulse response signal. The function $\epsilon(\cdot)$ is often modeled as a rapid rise-and-decay profile that captures the postsynaptic current evoked by an incoming spike. In laboratory settings, neurons can also receive artificially induced activity, which can be included in $I(t)$ as an additional external input. When postsynaptic currents are applied to a neuron, the membrane potential increases over time and, upon reaching a fixed threshold $\vartheta$, the neuron emits a spike, and the membrane potential resets to its resting value
\begin{equation}\label{eq:lif-spike}
    V(t) \geq \vartheta \implies V(t+\mathrm{d}t) = V_r \quad \text{and emit a spike.}
\end{equation}
Here, $\mathrm{d}t$ denotes an infinitesimal time step (or the simulation time step in discrete implementations). Note that, if there is no input current, \eqref{eq:lif-voltage} describes how the membrane potential $V(t)$ decays towards its resting value $V_{\text{r}}$ with a characteristic time constant $\tau_m$.
We will refer to spiking neural networks based on LIF neurons as LIF-SNN models.

\subsubsection{Synaptic Adaptation by Spike-Timing-Dependent Plasticity}\label{sec:stdp}

The synaptic weight $w_{ij}$ determines how much a spike from a pre-synaptic neuron $j$ influences the post-synaptic current in a receiving neuron $i$. Depending on the sign of $w_{ij}$, the connection can be excitatory (positive weight) or inhibitory (negative weight).
\begin{figure}[tb]
    \centering
    \includegraphics[width=0.8\linewidth]{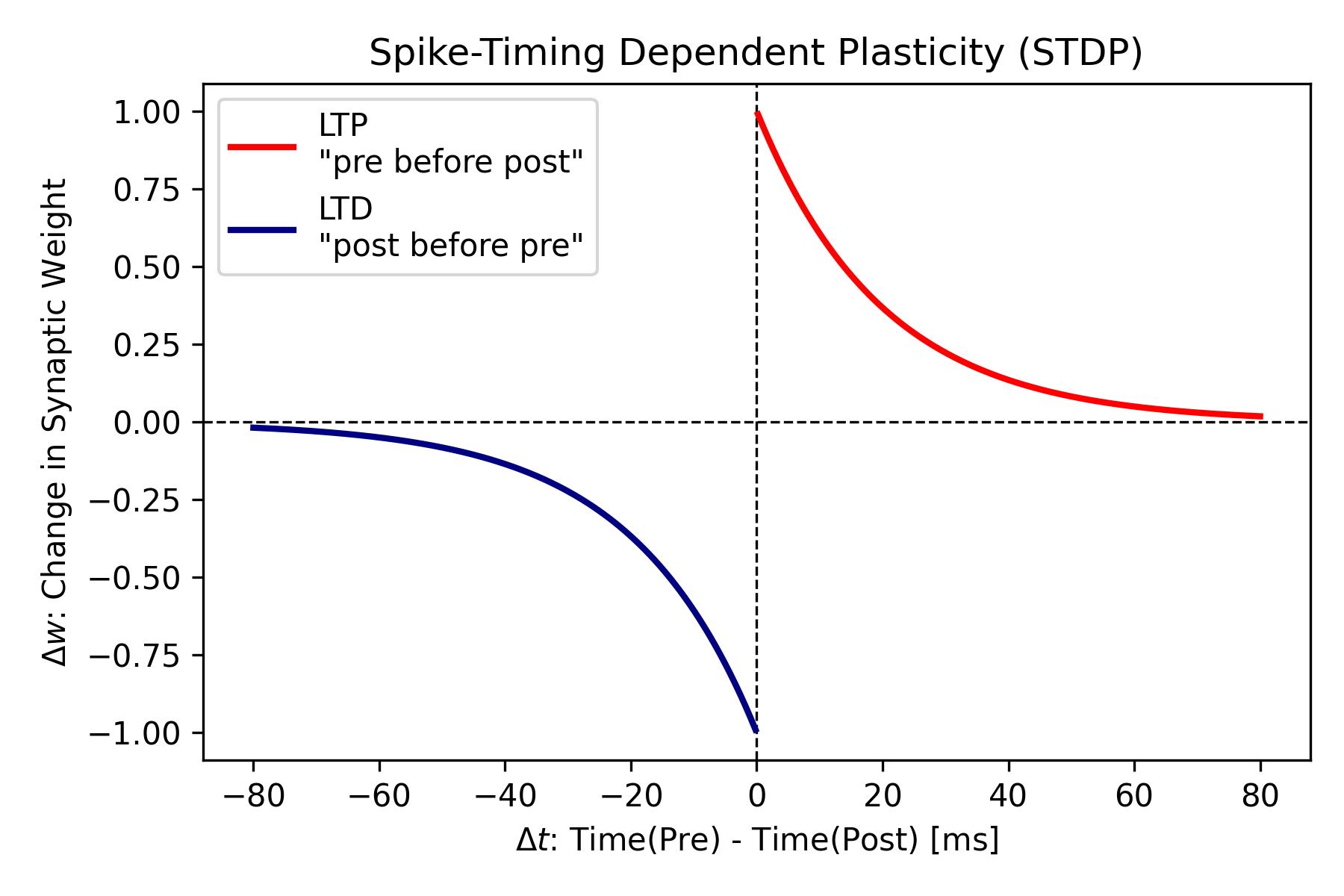}
    \caption{Synaptic weight updates for STDP, see Eq.\eqref{eq:STDP}.}
    \label{fig:stdp}
\end{figure}
Spike-Timing-Dependent Plasticity (STDP) is a biologically inspired learning rule that governs synaptic weight changes based on the relative timing of pre- and post-synaptic spikes ~\cite{song2000competitive}. It is a form of Hebbian learning, often summarized as "neurons that fire together, wire together", but with a temporal dependence that refines synaptic modifications.
If a presynaptic neuron fires before a postsynaptic neuron, the synapse is strengthened, a process known as long-term potentiation (LTP). Conversely, if a postsynaptic neuron fires before a presynaptic neuron, the synapse is weakened, known as long-term depression (LTD).
This mechanism is illustrated in Fig.~\ref{fig:stdp}, where the synaptic weight change $\Delta w_{ij}$ varies as an exponential function of the spike timing difference $\Delta t$. The plot shows the characteristic asymmetry of STDP, with potentiation for positive $\Delta t$ and depression for negative $\Delta t$. The STDP weight update rule is mathematically described as
\begin{equation}\label{eq:STDP}
\Delta w_{ij} = 
\begin{cases} 
A^+ e^{-\Delta t / \tau^+}, & \text{if } \Delta t > 0 \quad (\text{j-before-i, LTP}) \\
-A^- e^{\Delta t / \tau^-}, & \text{if } \Delta t < 0 \quad (\text{i-before-j, LTD})
\end{cases}
\end{equation}
where
\begin{itemize}
    \item \( \Delta w_{ij} \) represents the synaptic weight change,
    \item \( \Delta t = \hat{t}_{\text{post}} - \hat{t}_{\text{pre}} \) is the spike timing difference,
    \item \( A^+ \) and \( A^- \) are scaling factors for LTP and LTD, respectively,
    \item \( \tau^+ \), \( \tau^- \) are time constants governing the decay of potentiation and depression~\cite{song2000competitive}.
\end{itemize}

% \section{Spike-based Message Passing in a LIF-SNN}\label{section:solution}
\section{Proposed Framework}\label{section:solution}

In this section, we present our proposed solution: a framework for implementing Bayesian inference in linear Gaussian models through spike-based message passing in LIF-SNN models. First, we describe how Gaussian messages can be encoded in spike trains. We then detail the implementation of spike-based message update rules in LIF-SNNs for the linear node functions listed in Table~\ref{table:mp-n}.

\subsection{Representing Gaussian Messages by Spike Trains}\label{subsection:spikecoding}

The first step in implementing Gaussian message passing in SNNs is to represent Gaussian messages as spike-based signals. In computational neuroscience, spike coding schemes are typically classified into two categories: rate coding and temporal coding. Rate coding represents information as the average or instantaneous firing rate of individual neurons or populations. In contrast, temporal coding conveys information through the precise timing of each spike or the intervals between spikes \cite{gerstner2002spiking}.  Empirical findings indicate that biological neurons often utilize distinct coding strategies based on their functional roles and task demands~\cite{carleton2010coding}.

Population coding is a specialized form of rate coding in which the joint activity of spike rates across multiple neurons represents information. It captures essential features of biological neural coding while remaining simple enough for theoretical analysis~\cite{wu2002population}. This coding method is mathematically well-formulated and widely used in probabilistic simulations~\cite{zemel1998probabilistic, ma2006bayesian, sanger1996probability}. Population coding is also supported by biological evidence; for example, in the hippocampus of rats, the animal’s position is encoded through the combined activity of many place cells, each firing maximally at a different preferred location~\cite{o1971hippocampus}. Related studies have reported similar population-based representations in some other brain areas, including the prefrontal cortex and middle temporal visual area of macaques~\cite{ceccarelli2023static, maunsell1983functional}, as well as the frontal cortices of humans~\cite{shih2023electrophysiological}.

\subsubsection{Encoding Scheme}\label{sec:encoding} In the following, we consider sequences of time intervals, each of duration $T_s$. During an interval, the parameters of a Gaussian message  $\mu_x(x) = \mathcal{N}(x \mid m, \sigma^2)$ are assumed to remain fixed. For each time interval, we convert ("encode") the ``moment" parameters $\{m, \sigma^2\}$ to a set of ``firing rate" parameters $\{s_i,r_i\}_{i=1}^N$, to be defined below.

We adopt a population coding strategy, comprising a set of $N$ neurons to represent a Gaussian message $\mu_x(x) = \mathcal{N}(x \mid m, \sigma^2)$ for a variable $x \in \mathbb{R}$. Each neuron's firing rate $r_i$  ($i=1,2,\ldots,N$) can be interpreted as the average number of spikes in the time interval $T_s$. 
To construct the firing rate code, we assign to each neuron $i$ a unique spatial location $s_i$, spaced evenly across the interval $[m - 3\sigma_,\, m + 3\sigma]$, by
\begin{equation}
    s_i = m -3\sigma + (i-1)\cdot \frac{6\sigma}{N-1} \quad \text{for }i=1,2,\ldots,N\,.
\label{eq:rate-locations}
\end{equation}

\begin{figure}[tb]
\centering 
\includegraphics[width=1\columnwidth]{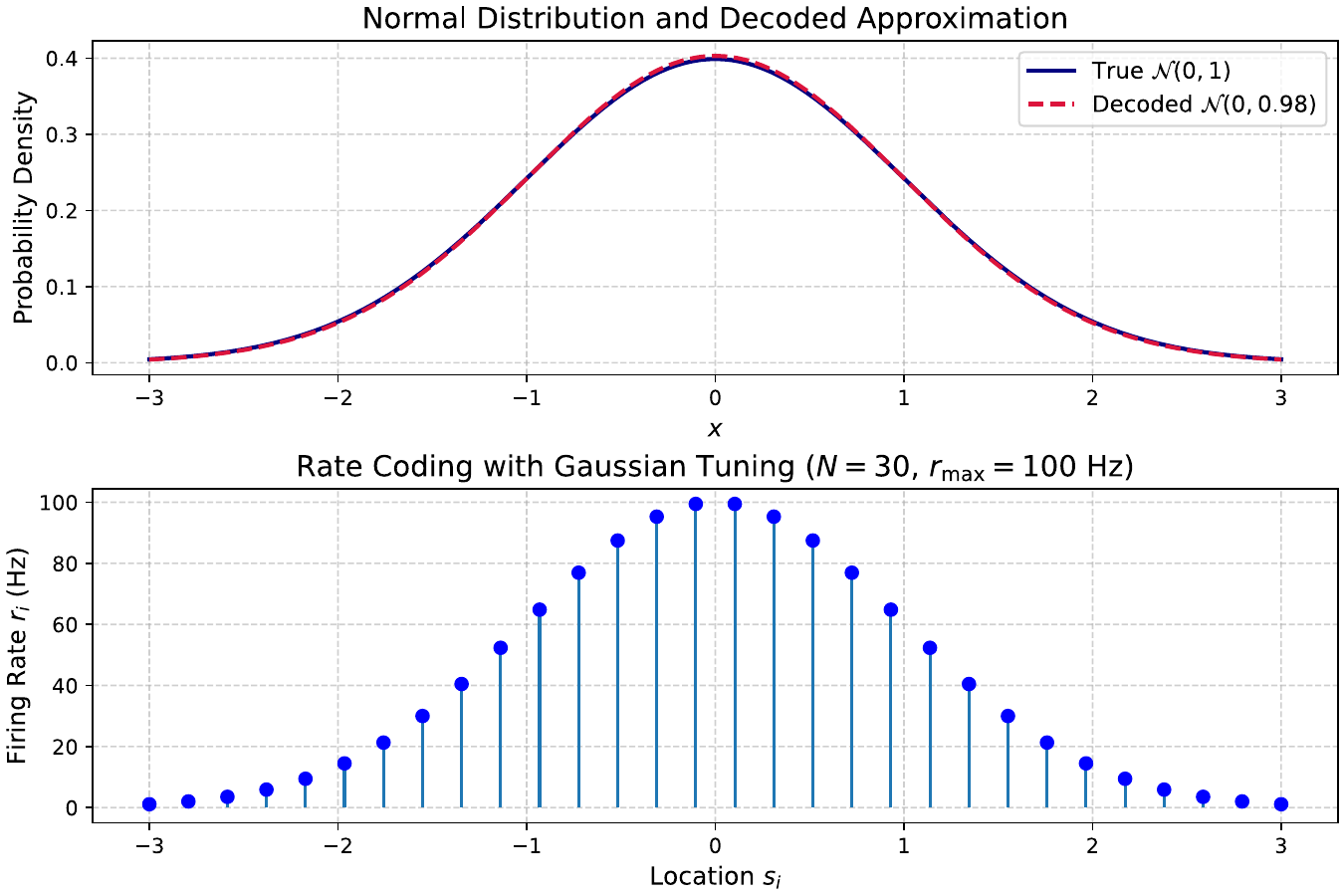}
\caption{Population coding for Gaussian messages using \eqref{eq:rate-locations}, \eqref{eq:firing_rate} for encoding and \eqref{eq:decod} for decoding. 
Top: The standard normal distribution $ \mathcal{N}(0, 1) $ representing a Gaussian message, along with the decoded message. (The decoded message is barely visible because the original and decoded messages largely overlap, as desired). Bottom: Firing rates of $ N = 30 $ neurons, $r_{\text{max}} = 100$ [Hz]. }
% \bdv{Please change x-axis legend to: location $s_i$ and replace "Preferred Stimulus" by "Location".}}
\label{fig:normal_encoding}
\end{figure}

This range is chosen empirically to sufficiently cover the effective support of the Gaussian (the Gaussian integral from $m-3\sigma$ to $m+3\sigma$ equals $0.9973$), while maintaining computational efficiency.
The firing rate $r_i$ of the $i^{th}$ neuron is determined to be the value of a Gaussian function at spatial location $s_i$ scaled by a maximum firing rate $r_{\text{max}}$. This means the firing rate is proportional to the probability under a Gaussian density function centered at $m$ with variance $\sigma^2$,
\begin{subequations}
    \begin{align}
        r_i &= r_{\text{max}} \cdot \exp\left(-\frac{(s_i - m)^2}{2\sigma^2}\right) \label{eq:firing_rate} \\ 
        &\propto \mathcal{N}\big(s_i \given m, \sigma^2 \big) \, .
    \end{align}    
\end{subequations}
Similar approaches have been used in previous studies to represent continuous variables through population codes~\cite{zemel1998probabilistic, ma2006bayesian}. As an example, Fig.~\ref{fig:normal_encoding} illustrates this encoding scheme for a Gaussian message $\mu_x(x) = \mathcal{N}(x \mid 0, 1)$.

\subsubsection{Decoding Scheme}\label{sec:decoding}
To decode the spike-based representation back into the moment parameters of the Gaussian message, we utilize the location values $s_i$ and their corresponding firing rates $r_i$. The decoded mean $\hat{m}$ and variance $\hat{\sigma}^2$ are computed as the weighted mean and variance of the rate values,

\begin{align} \label{eq:decod}
    \hat{m} = \frac{\sum_{i=1}^{N} r_i s_i}{\sum_{i=1}^{N} r_i} \, , \qquad 
    \hat{\sigma}^2 = \frac{\sum_{i=1}^{N} r_i (s_i - \hat{m})^2}{\sum_{i=1}^{N} r_i} \, .
\end{align}    

In summary, a Gaussian message can be represented either by its moments, $\{m, \sigma^2\}$, or by its firing rate representation, $\{s_i, r_i\}_{i=1}^N$. Going forward, we use the term ``standard message passing`` when Gaussian messages are expressed in terms of their moment parameters, and ``spike-based message passing`` when they are expressed in terms of firing-rate parameters.

% \subsection{SNN-Based Simulation of the Sum-Product Update Rule for Linear Factor Nodes}
\subsection{Message Passing-based Inference in LIF-SNN models }

In this section, we address how to construct LIF-SNN models that enable spike-based message passing in factor graphs, allowing spike-encoded Gaussian messages to be processed in the same way as in standard FFG nodes. In the following three sections, we present the LIF-SNN models for the equality, addition, and multiplication nodes listed in Table~\ref{table:mp-n}.

\subsubsection{Equality Node}\label{sec:equality-node}
\begin{figure}
    \centering
    \includegraphics[width=0.9\linewidth]{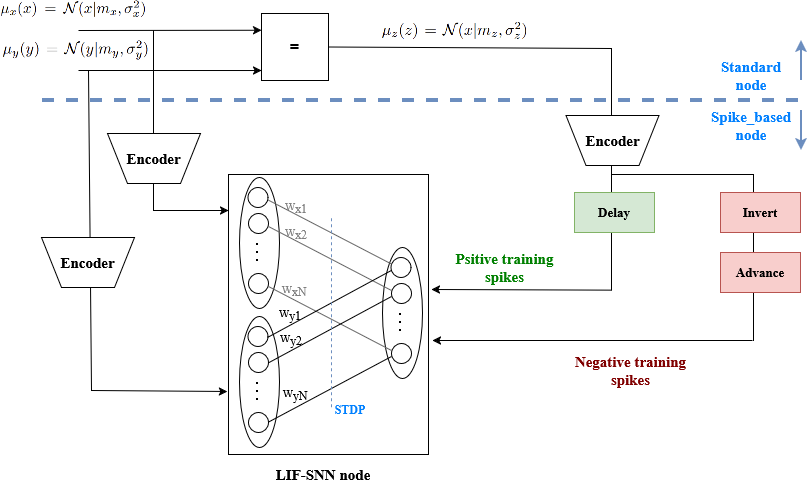}
    \caption{Training-phase architecture of the LIF-SNN node for the equality operation. The input distributions are first used to obtain the numerical ground-truth output. These distributions are then encoded into spike trains. Positive training spikes are generated by applying a time delay to the encoded spikes, while negative training spikes are produced by applying a time advance to the flipped spikes. See Section~\ref{sec:equality-node} for details of the STDP-based training process.}
    \label{fig:equality}
\end{figure}

We trained a LIF-SNN model to perform sum-product message passing of spike-encoded Gaussian messages. Consider the architecture of the training session as displayed in Fig.~\ref{fig:equality} for the following description.
There are two nodes in the system. The top node is a standard equality node $f(x,y,z) = \delta(z-x)\delta(z-y)$. The other node, at the bottom, is a spike-based LIF-SNN model. The input layer of the LIF-SNN model consists of $N=100$ LIF neurons for $x$ and $100$ LIF neurons for $y$. The output of the LIF-SNN model is a layer of $100$ LIF neurons that encode outgoing messages for $z$. Messages from each neuron for $x$ and for $y$ are weighted and connected to the corresponding output layer neuron in $z$. The total LIF-SNN model contains $200$ weights ($100$ from $x$ to $z$ and $100$ from $y$ to $z$). The weights are trainable by STDP.

First, a training set of standard-encoded Gaussian messages $\mu_x(x) = \mathcal{N}(x\mid m_x,\sigma_x^2)$ and $\mu_y(y) = \mathcal{N}(y\mid m_y,\sigma_y^2)$ were generated. These messages were passed through the standard equality node that generated outgoing messages $\mu_z(z) = \mathcal{N}(x\mid m_z,\sigma_z^2)$, see \eqref{eq:eq-node-forward} and Table~\ref{table:mp-n} for details. 
Next, each of the messages $\mu_x(x)$, $\mu_y(y)$, and $\mu_z(z)$ was converted into its firing rate representation, following the procedure described in Section~\ref{sec:encoding}. These rate-encoded messages were then presented to their corresponding layers in the LIF-SNN model.
The $\mu_z(z)$ messages were delayed by $\Delta t$ to serve as teacher-forcing signals in an LTP-STDP learning process, see Eq.~\eqref{eq:STDP}. In addition, $\mu_z(z)$ was inverted and temporally advanced by $\Delta t$ to drive an LTD--STDP learning process. This STDP training architecture was inspired by~\cite{mo2021logicsnn}. 
% The parameters for the STDP learning process were $A^+=0.25$, $A^- = -0.125$, $\tau^+=20$ ms, $\tau^-=20$ ms. 
%
%
%
\begin{figure}[htbp]
    \centering
    \includegraphics[width=\textwidth]{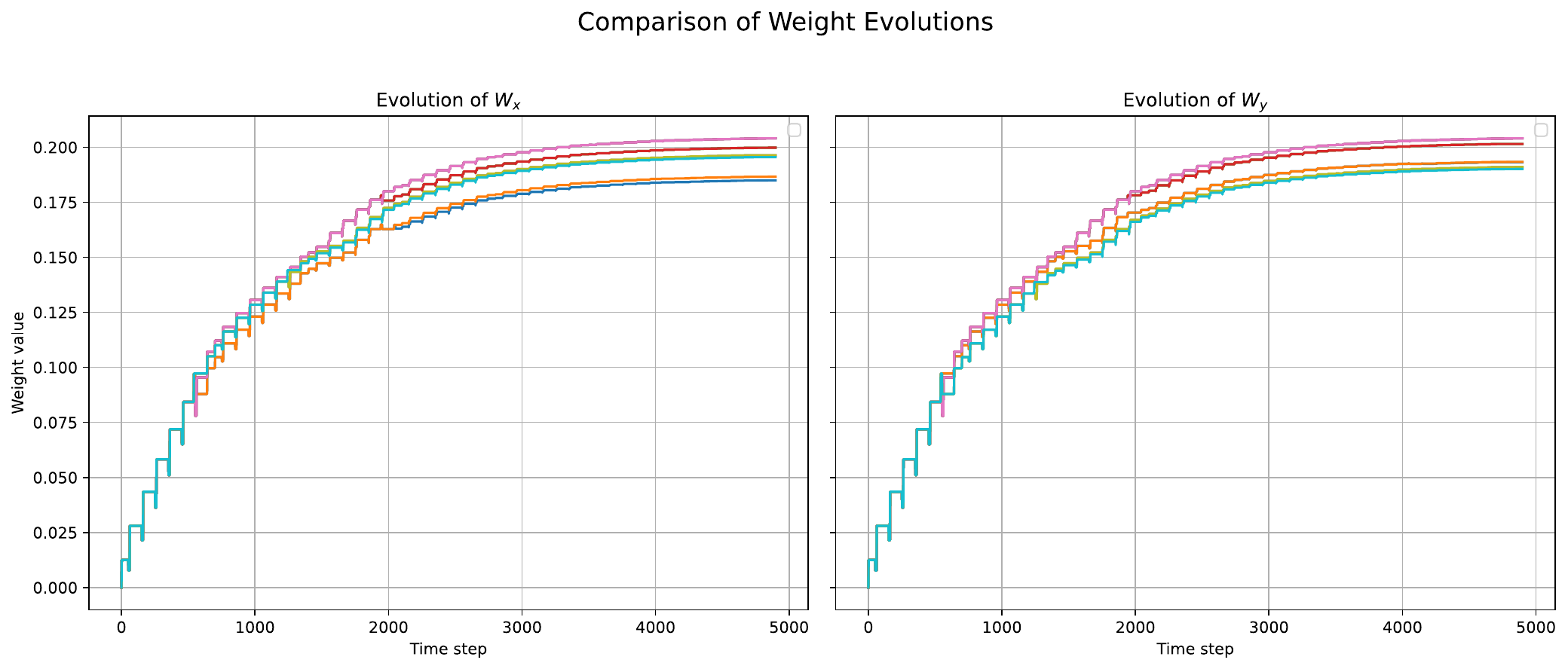}
    \caption{Comparison of the evolution of synaptic weights $W_x$ and $W_y$ during training of the equality node function using STDP. The weights $W_x$ and $W_y$ connect input spikes corresponding to $\mu_x$ and $\mu_y$, respectively, to the output layer.}
    \label{fig:weight_evolution}
\end{figure}
In Fig.~\ref{fig:weight_evolution}, we show the evolution during training of five randomly selected weights from $x$ to $z$, as well as five weights from $y$ to $z$. The parameter set used is detailed in Table~\ref{tab:parameters}. The results indicate that the weights converge to stable endpoints.
\begin{figure}
    \centering
    \includegraphics[width=0.9\linewidth]{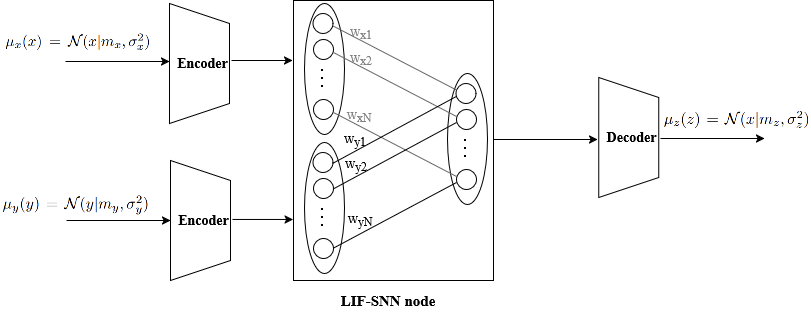}
    \caption{Testing-phase architecture of the LIF-SNN node for the equality operation. After training, the synaptic weights of the LIF-SNN node are fixed. The input distributions are encoded into spike trains and fed into the trained network. A decoder is then used to reconstruct the output distribution from the resulting spike activity.}
    \label{fig:equality_test}
\end{figure}

\begin{figure}[tb]
\centering
\includegraphics[width=1\columnwidth]{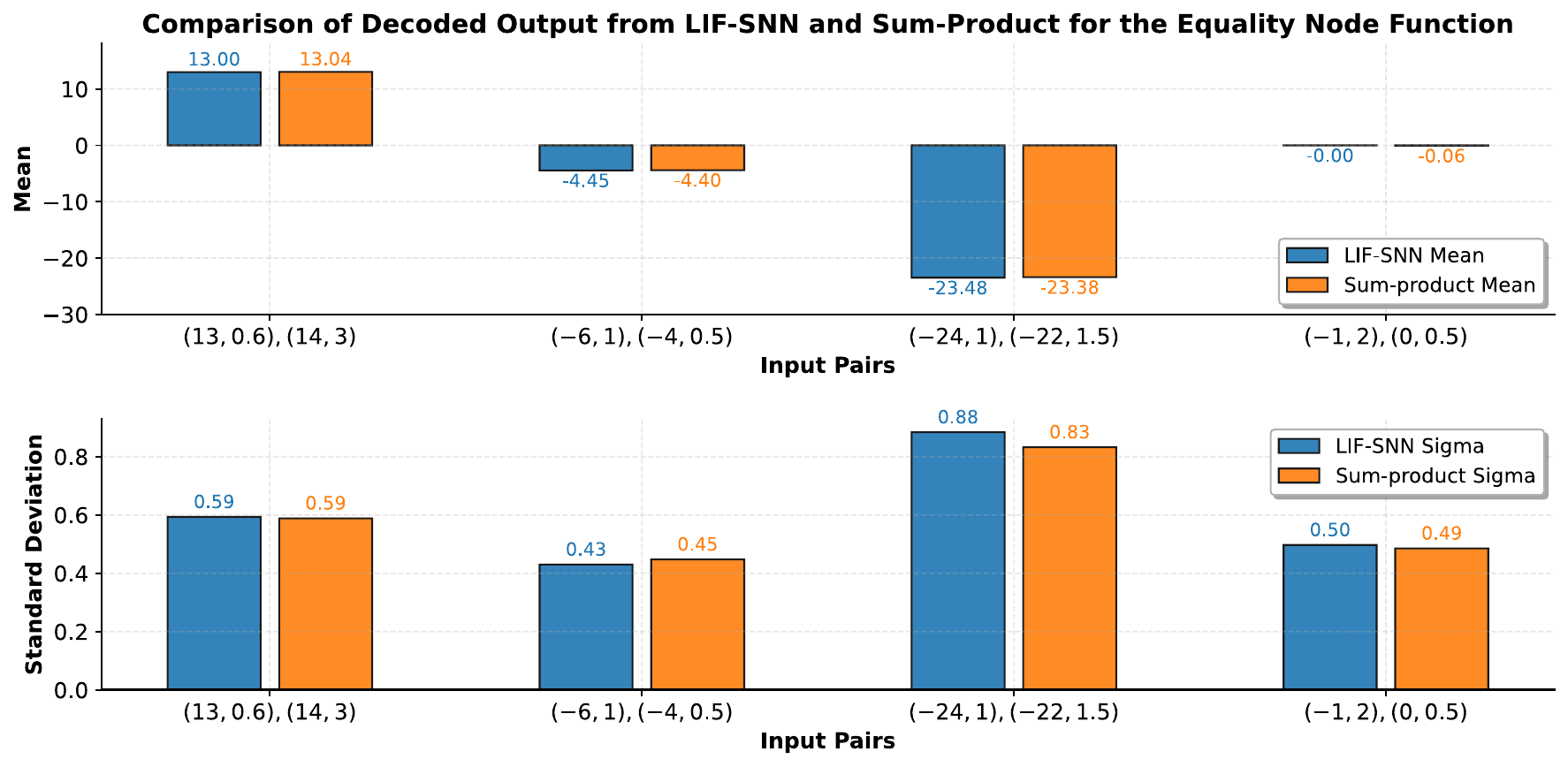}
\caption{Comparison of the resulting mean and standard deviation produced by the LIF-SNN model and the sum-product algorithm for the equality operation. }%\bdv{This figure seems inappropriate. Why are there connections between the different testing samples? We need a bar graph here.}}
\label{fig:normal_mp}
\end{figure}

After the training, we replace the training spikes with a decoder as shown in Fig~\ref{fig:equality_test}. To assess the performance of the trained LIF-SNN model, we compared $\mu_z(z)$ on a testing data set. The results, shown in Fig.~\ref{fig:normal_mp}, confirm that the trained LIF-SNN model behaves correctly as an equality node for spike-encoded Gaussian messages. 
Since the equality node is symmetric across $x$, $y$, and $z$, that is, any re-labeling of the variables yields the same message update rules, we use the same trained LIF-SNN to generate the backward messages.

\subsubsection{Addition Node}\label{sec:addition-node} 

We begin this section by exploring the summation node function in greater depth. Considering the delta function $ \delta(z - (x + y)) $ for the random variables $x$, $y$, and $z$, the output message according to equation~\eqref{eq:sum-product}, can be obtained by

\begin{equation}
\overrightarrow{\mu}_z(z) = \int \overrightarrow{\mu}_{x}(x)\overrightarrow{\mu}_{y}(y) \delta(z - (x + y)) \, dx \, dy.
\end{equation}
By applying the sifting property of the Dirac delta function, this expression simplifies to
\begin{equation}
\overrightarrow{\mu}_z(z) = \int \overrightarrow{\mu}_{x}(x) \overrightarrow{\mu}_y(z - x) \, dx,
\end{equation}
which is the definition of the convolution of $\mu_{x}$ and $\mu_{y}$ \cite[Sec.~5]{loeliger2004introduction},
\begin{equation}
\overrightarrow{\mu}_z(z) = (\overrightarrow{\mu}_{x} * \overrightarrow{\mu}_{y})(z).
\end{equation}
This section proposes a novel Convolutional Spiking Neural Networks (CSNNs) for simulating summation node function.

CSNNs have been employed in various domains, including capturing spatial and temporal relationships in event-based sequences~\cite{xing2020new}, detecting anticipatory braking intentions from EEG signals~\cite{lutes2024convolutional}, and performing speech recognition~\cite{pellegrini2021low}.

\begin{figure}
    \centering
     \includegraphics[width=1\linewidth]{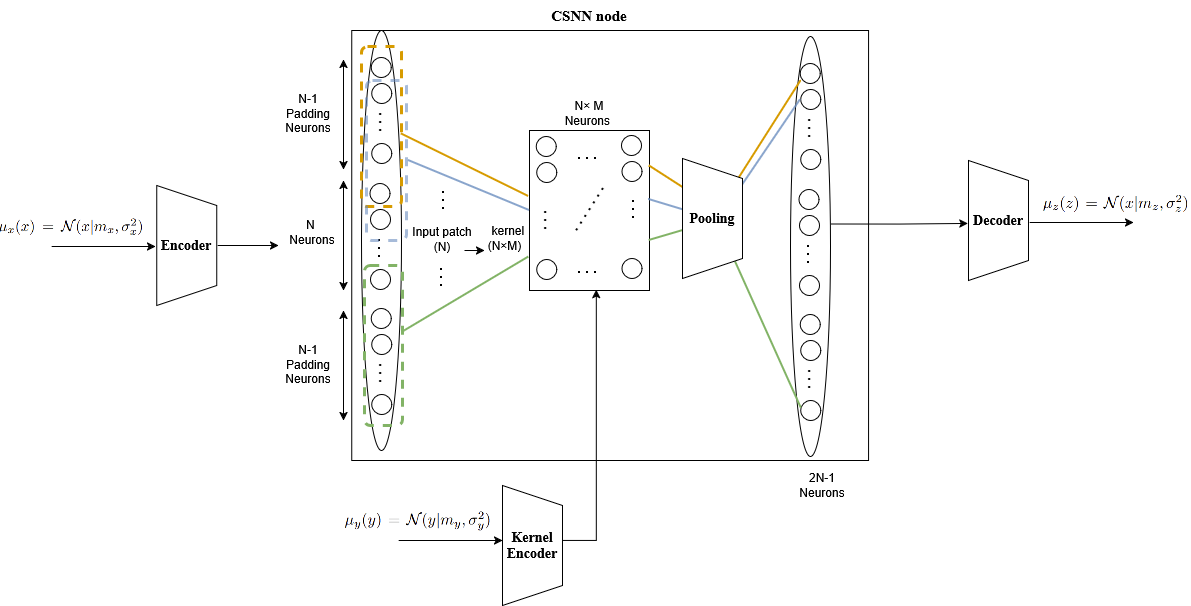}
    \caption{Illustration of the proposed convolution network for the summation node. The kernel consists of a two-dimensional group of neurons, with their thresholds set according to one of the input Gaussian messages. The kernel slides across another group of neurons representing the other Gaussian input, where fully connected synapses transmit the spikes. These spikes shape the output message.}
    \label{fig:conv}
\end{figure}

\begin{figure}
    \centering
     \includegraphics[width=0.85\linewidth]{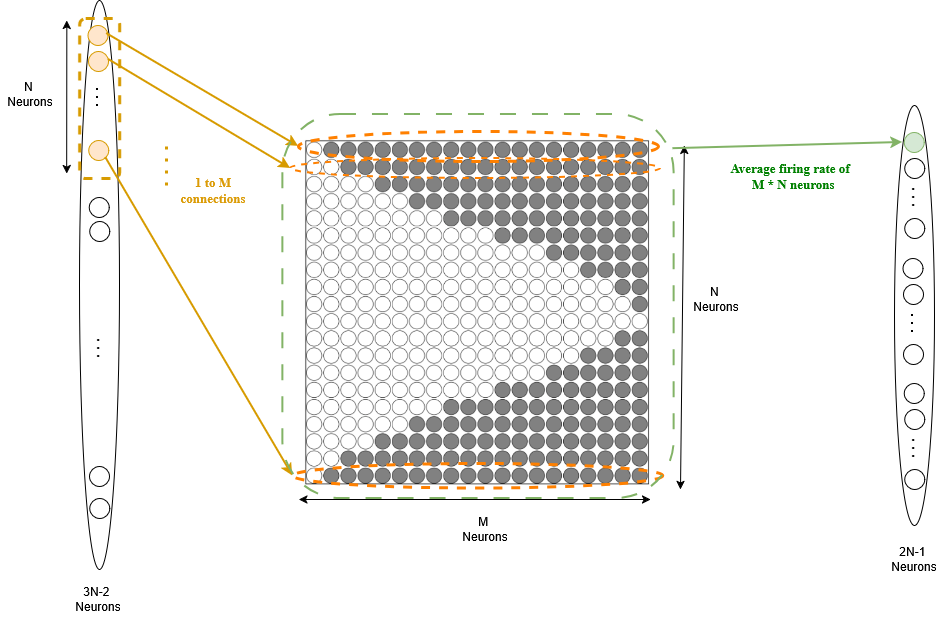}
    \caption{Schematic illustration of the kernel threshold connectivity. Each of the N input neurons on a patch (left)  projects one-to-M connections into a group of $M$ neurons of the kernel (center). The color pattern encodes the threshold distribution generated by a Gaussian input message. White circles indicate neurons with low thresholds that readily pass incoming spikes, whereas gray circles denote neurons with high thresholds that inhibit spike transmission.  The N × M average firing rate is then mapped onto the firing rate of a single neuron in the output layer (right).}
    \label{fig:conv2}
\end{figure}

In our proposed model, the two input messages are convolved with each other by encoding one message to make the input layer and the other to form the convolution kernel. 
The input layer consists of $N$ neurons whose firing rates follow the encoding scheme described in Section~\ref{sec:encoding}, along with an additional $2N - 2$ neurons with zero firing rates that are padded as illustrated in Fig.~\ref{fig:conv}. 

The kernel is encoded to represent the message $\overrightarrow{\mu}_{y}$, more precisely, the thresholds of the kernel's neurons are set based on moments of the input message. This kernel is composed of $N$ groups, each with $M$ LIF neurons. Each group shares the same position $s_{y,i}$, which is computed in the same way as defined in Eq~\eqref{eq:rate-locations}.  
Within each group, a subset of LIF neurons is assigned a low firing threshold, $\vartheta_l$, allowing them to be activated by spike activity in the previous layer, while the remaining neurons are assigned a high firing threshold, $\vartheta_h$, suppressing their activation. 
The number of neurons with low and high thresholds within each group of size $M$ is determined as follows
\begin{equation}
        M_{\text{l}} = \left\lfloor 
        \exp\!\left(-\dfrac{\left| s_i - m_y \right|^2}{2\sigma_y^2}\right) 
        \right\rfloor \times M, \\ \qquad
        M_{\text{h}} = M - M_{\text{l}}.
\end{equation}
In Fig.~\ref{fig:conv2}, an example of such a kernel is illustrated. In this example, $M=N=20$ were chosen for illustrative purposes.

To perform the convolution, input patches (or segments) of length  $N$ are extracted from the input layer. 
The indices corresponding to these patches are defined as
\begin{equation}
    \mathcal{P}_i=\{n_{i},n_{i+1},\dots,n_{i+N-1}\} \quad \text{for } i=1,\dots,2N-1.
\end{equation}
where $ \mathcal{P}_i $ denotes the $ i $-th patch, and $n_j $ represents the indices of the input-layer neurons that constitute the $ i $-th patch. 
This procedure yields $ 2N - 1 $ patches, each connected to the convolution kernel. 
For each connection, every neuron within a patch is connected by $M$ synapses to its corresponding group in the kernel. 
The synaptic weights for these connections are fixed and configured so that the activation of each patch neuron drives its corresponding group neurons with low thresholds to fire spikes. 
The weight is set to
\begin{equation}
   w = \vartheta_{\text{l}} - V_{\text{rest}} + \epsilon, 
\end{equation}
where $\epsilon$ is a small positive constant. This connection pattern for a single patch is illustrated in Fig.~\ref{fig:conv2}.
Finally, the neurons in the kernel are connected to a pooling layer, which computes the average firing rate across all $N\times M$ neurons and transmits it to a single neuron in the output layer. 
Therefore, the number of neurons in the output layer is equal to the number of patches, i.e., $2N - 1$.

To decode the spike activity from the CSNN into a Gaussian distribution, a decoder identical to that described in Section~\ref{sec:decoding} is used. We define the spatial locations \( s_{z,i} \) as, 
\begin{equation}
    s_{z,i} = 2\,s_{\min} + \frac{i - 1}{2N - 2} \left( 2\,s_{\max} - 2\,s_{\min} \right),
\quad i = 1, 2, \dots, 2N - 1.
\end{equation}
where 
\begin{equation}
    s_{\min} = \min \bigl( \min_i s_{x,i}, \, \min_i s_{y,i} \bigr),
    \quad
    s_{\max} = \max \bigl( \max_i s_{x,i}, \, \max_i s_{y,i} \bigr).
\end{equation}

\begin{figure}[tb]
\centering
\includegraphics[width=1\columnwidth]{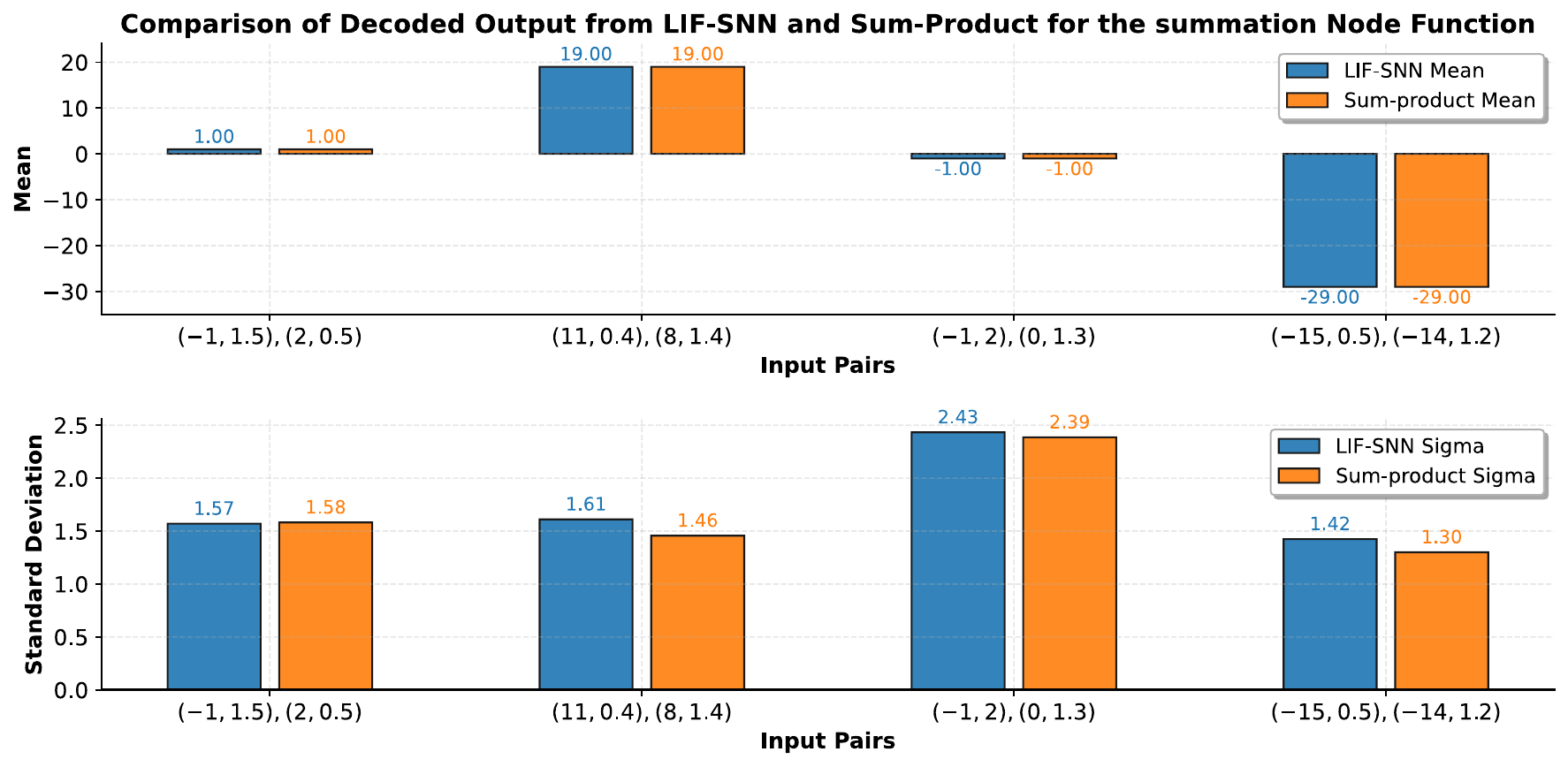}
\caption{Comparison of the resulting mean and standard deviation produced by the LIF-SNN model and the sum-product algorithm for the summation operation.}% \bdv{bar graph.}}
\label{fig:sum}
\end{figure}

Figure~\ref{fig:sum} presents a comparison between the decoded results produced by the proposed CSSNs and the numerical sum-product results for four representative pairs of input examples. The selected examples include negative values, relatively small numbers, and decimal quantities, illustrating the close agreement between the two methods.

We can adopt a similar structure to derive the backward message. Suppose we aim to compute the backward message  
$\overleftarrow{\mu}_y(y) = \mathcal{N}(y \mid m_y, \sigma_y^2),$
given the inputs $\overrightarrow{\mu}_x(x) = \mathcal{N}(x \mid m_x, \sigma_x^2)$ and $\overleftarrow{\mu}_z(z) = \mathcal{N}(z \mid m_z, \sigma_z^2)$. According to the sum--product computation summarized in Table~\ref{table:mp-n}, the encoding can be obtained by substituting $-m_x$ for $m_x$, while all other computational steps remain unchanged.

\begin{figure}[tb]
\centering
\includegraphics[width=1\columnwidth]{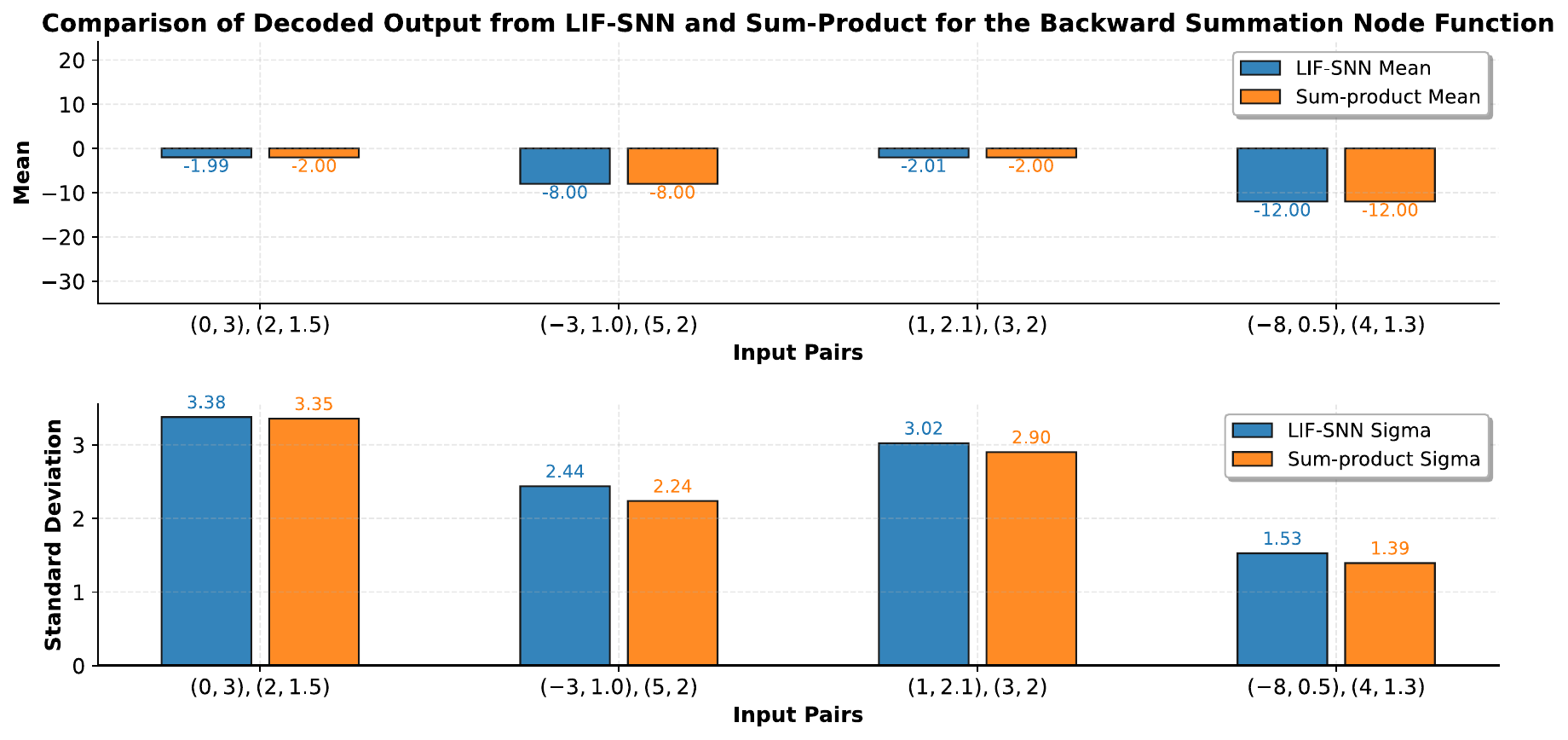}
\caption{Comparison of the resulting mean and standard deviation produced by the LIF-SNN model and the sum-product algorithm for the summation operation in the backward direction.}% \bdv{change to bar graph.}}
\label{fig:sum_bwd}
\end{figure}

The results on a representative test set are shown in Figure~\ref{fig:sum_bwd}, along with a comparison against the numerical sum--product results. The computation of the backward message on $x$, given $\overrightarrow{\mu}_y(y)$ and $\overleftarrow{\mu}_z(z)$, follows an analogous procedure.

\subsubsection{Multiplication Node}\label{sec:multiplication-node}

The type of multiplication considered in this section is the multiplication of a scalar with a Gaussian input message. It should be noted that the multiplication of two Gaussian input messages does not yield a Gaussian output message and therefore lies outside the scope of this paper. The corresponding Delta function for this operation is defined as $f_{\times}(x, y, z) = \delta(z - a y)$, which is also known as a \textit{scaling node}. We observe that this function preserves the neurons’ firing rates from input in the output while only shifting their spatial locations $s_i$.

\begin{figure}[tb]
\centering
\includegraphics[width=1\columnwidth]{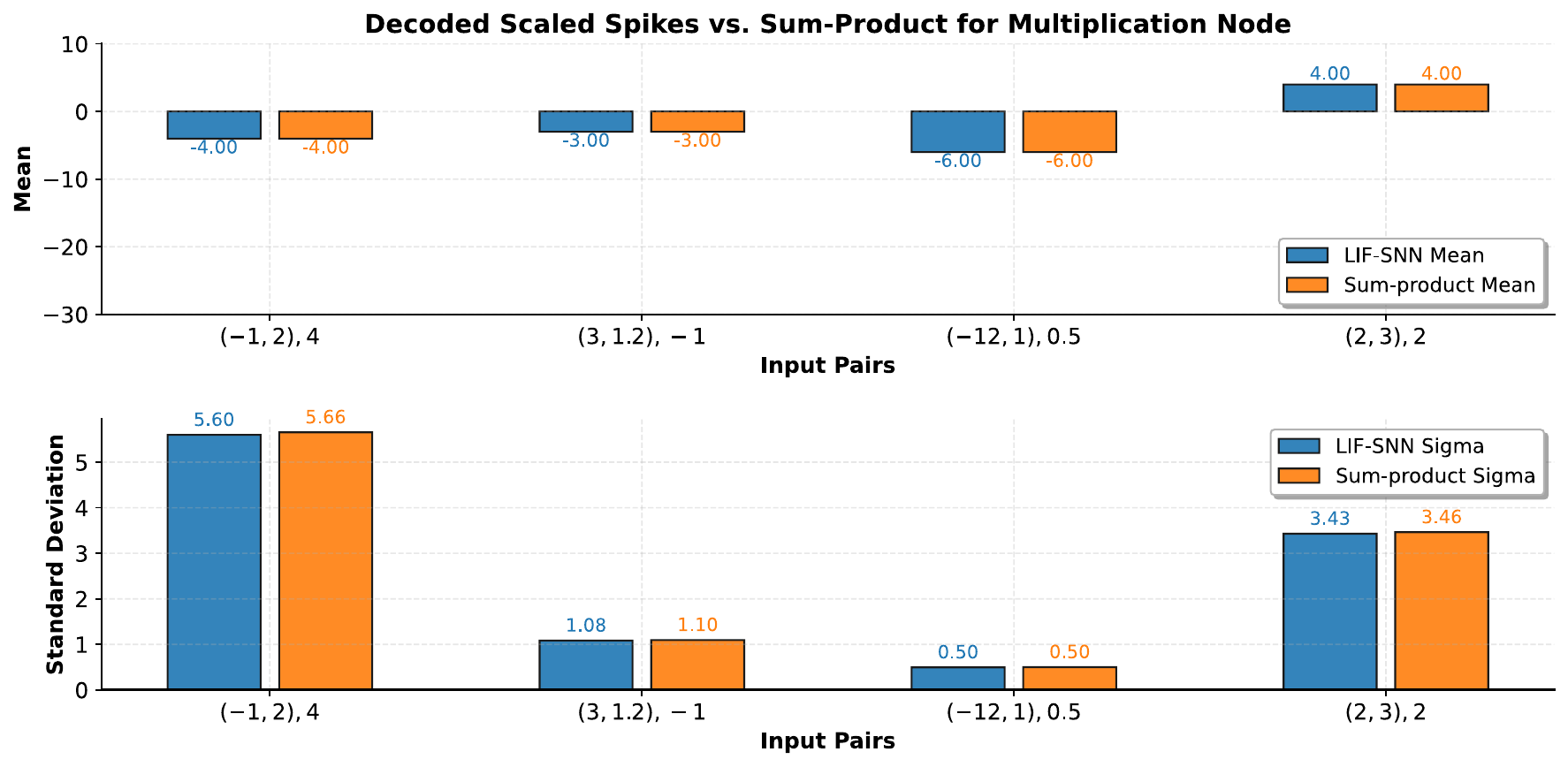}
\caption{Comparison of the decoded mean and standard deviation produced by scaling the position labels and the sum-product algorithm for the multiplication operation.} 
\label{fig:mult}
\end{figure}

\begin{figure}[tb]
\centering
\includegraphics[width=1\columnwidth]{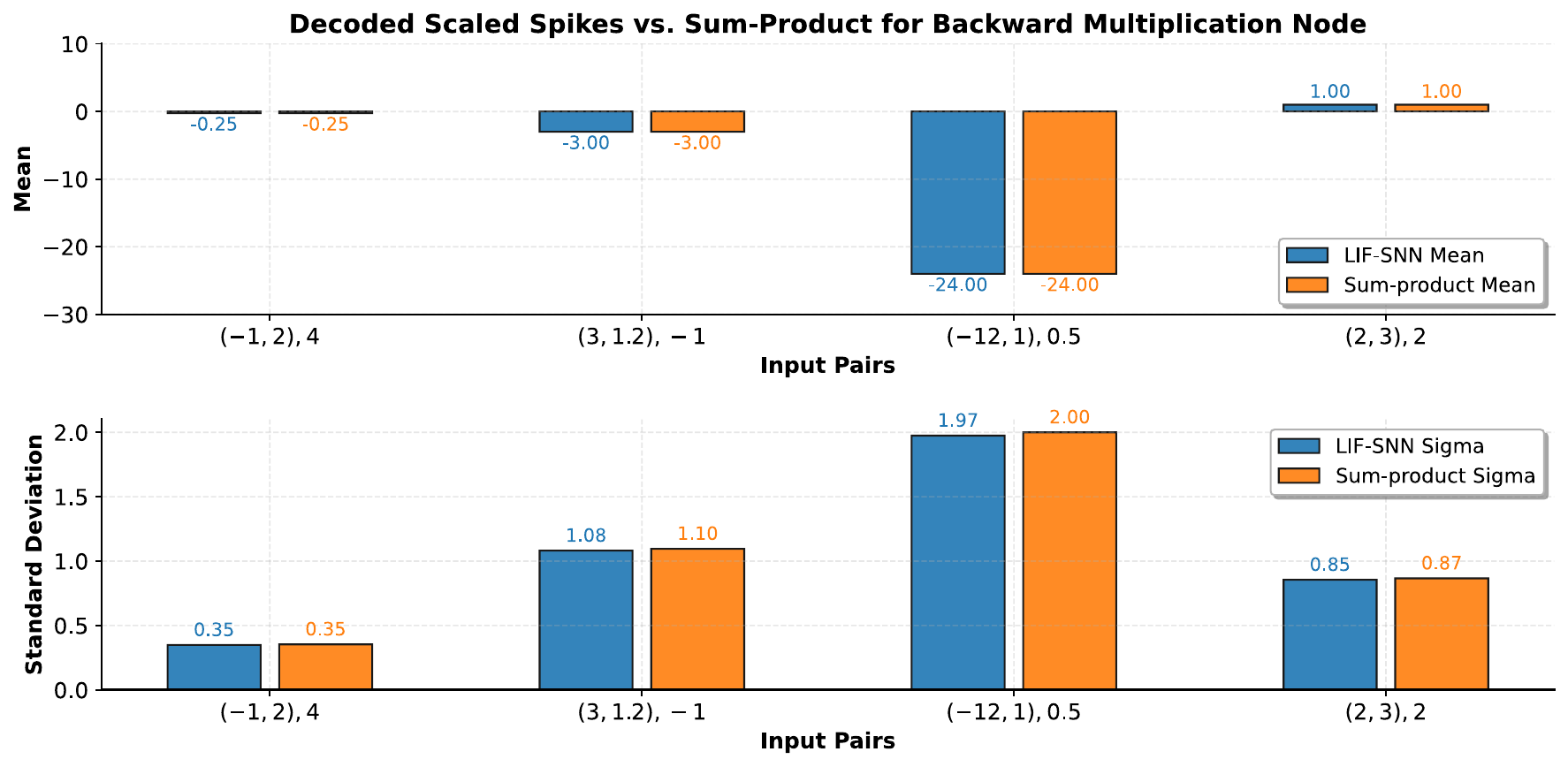}
\caption{Comparison of the decoded mean and standard deviation produced by scaling the position labels and the sum-product algorithm for the backward multiplication operation. }
\label{fig:mult_bwrd}
\end{figure}

The location updates can be expressed as  
\begin{equation}
    s_{z,i} = s_{y,i} + a.
\end{equation}
Figure~\ref{fig:mult} compares the results of the standard sum–product computation with those obtained from the LIF–SNN model by decoding the messages using the updated locations.  
The backward message is implemented analogously, with the location update defined as  
\begin{equation}
    s_{y,i} = s_{z,i} - a.
\end{equation}
The corresponding comparison is shown in Figure~\ref{fig:mult_bwrd}.

\section{Examples}\label{section:validation}
% \wmk{Rename this section to Experiments}
In this section, we demonstrate how two classical Bayesian algorithms,  Bayesian linear regression and the Kalman filter, can be implemented using spike-based message passing in LIF-SNN networks. In this section, we refer to SNN-based message passing as the process of encoding the input Gaussian messages, propagating them through the networks of LIF-SNN models described in the previous section, and decoding them back into the parameters of Gaussian distributions. The FFG, used to represent the message flow in SNN-based message passing, is analogous to the standard message-passing framework.

In this paper, we used the \texttt{Brian2} simulator~\cite{goodman2008brian} to simulate LIF-SNN models\footnote{The code will be made publicly available in the final version of the paper.}, which provides a flexible and efficient platform for SNN simulation. The parameter values used in this paper are listed in Table~\ref{tab:parameters}.

\begin{table}[h!]
\setlength{\tabcolsep}{2.2pt}
\renewcommand{\arraystretch}{1.5}
\centering
\caption{Parameters LIF-SNN ($N=100$, $M=50$) and STDP used in examples.}
\label{tab:parameters}
\begin{tabular}{|c|c|c|c|c|c|c|c|c|c|c|c|c|c|c|}
\hline
 $\vartheta$ & $V_r$ & $R_m$ &  $A^+$ & $A^-$ & $\vartheta_l$ & $\vartheta_h$ & $r_{\text{max}}$ & $T_s$ &
$w_{\max}$ & $w_{\min}$ & $\tau^+$ & $\tau^-$ & $\tau_m$ & $\Delta t$  \\
 \hline 
 -50 mV & -80 mV &  1 &  0.25 & -0.125  & -0.50 & -0.30 & 100 & 1 s &
1 & -1 & 20 ms & 20 ms & 10 ms & 1 ms  \\
\hline
\end{tabular}
\end{table}

\subsection{Kalman Filtering}\label{sec:KF-example}

State estimation in dynamic systems is a central problem in control theory, robotics, and signal processing. In this subsection, we address the car tracking problem introduced in Section~\ref{sec:background}.

Based on \ref{eq:KF_model1}, the generative model in \ref{eq:KF_model}, and the corresponding FFG in \ref{fig:FFG-KF}, our goal is to track $p(x_t\mid y_{1:t})$, i.e., to estimate the position $x_t$ given the past observations. The Kalman filter is a recursive solution to this problem, in the sense that $p(x_t\mid y_{1:t})$ is updated from a given previous estimate $p(x_{t-1}\mid y_{1:t-1})$ and a new observation $y_t$. Assuming a given previous estimate 
$$
p(x_{t-1}\mid y_{1:t-1}) = \mathcal{N}(x_{t-1}\mid m_{t-1},\sigma_{t-1}^2) \,,
$$ 
Kalman filtering can be executed by evaluating the following set of equations at time step $t$ \cite{sarkka2023bayesian}:
\begin{subequations}\label{eq:KF-closed-form}
 \begin{align}
   \bar{m}_t &= m_{t-1} + m_u \\
    \bar{\sigma}_t^2 &= \sigma_{t-1}^2 + \sigma_x^2 + \sigma_u^2  \\
    k_t &= \frac{\bar{\sigma}_t^2}{\bar{\sigma}_t^2 + \sigma_y^2} \quad \text{(Kalman gain)} \\
    m_t &= \bar{m}_t + k_t\cdot(y_t-\bar{m_t})\\
    \sigma_t^2 &= (1-k_t)\bar{\sigma}_t^2.
\end{align}   
\end{subequations}

Alternatively, Kalman filtering can be executed by sum-product message passing on the factor graph of the generative model \cite{loeliger_factor_2007}. 

\begin{table}[h!]
\centering
\setlength{\tabcolsep}{9pt}
\renewcommand{\arraystretch}{1.4}
\caption{Comparison of Message Passing and Kalman Gain Method}
\label{tab:kalamn-result}
\begin{tabular}{|c|c|c|c|c|}
\hline
\textbf{Step} & \textbf{Quantity} 
& \makecell{Kalman Gain\\Method} 
& \makecell{Message\\Passing} 
& \makecell{SNN Message\\Passing} \\
\hline
\multirow{3}{*}{1} 
 & Kalman Gain ($k_1$)             & 0.336 & ---  & ---\\
 & Posterior mean ($m_1$)        & 2.993 & 2.993 & 3.029\\
 & Posterior variance ($\sigma_1^2$)      & 0.671 & 0.671 & 0.654\\
\hline
\multirow{3}{*}{2} 
 & Kalman Gain ($k_2$)             & 0.254 & --- &  ---\\
 & Posterior mean ($m_2$)        & 5.434 & 5.434 & 5.408\\
 & Posterior variance ($\sigma_2^2$)      & 0.508 & 0.508 & 0.484\\
\hline
\multirow{3}{*}{3} 
 & Kalman Gain ($k_3$)             & 0.206 & --- &  ---\\
 & Posterior mean ($m_3$)     & 7.860& 7.860 & 7.853\\
 & Posterior variance ($\sigma_3^2$)   &  0.411 &  0.411 &  0.382\\
\hline
\multicolumn{5}{|c|}{\vdots} \\
\hline
\multirow{2}{*}{10} 
  & Prediction mean ($m_{10}$)     & 24.419 & 24.419 & 24.804\\
 & Prediction variance ($\sigma_{10}^2$)   &  0.222 & 0.222 &  0.113\\
\hline
\end{tabular}
\end{table}
In the experiments described below, we used the following parameter settings. $m_u = 4$, $\sigma_u^2=0.01$, $\sigma_y^2 = 2$. The initial position was set to $p(x_0) = \mathcal{N}(x_0|1.0,0.5)$.The observation sequence was generated according to the stochastic process 
\begin{equation}
    y_t = m_u + y_{t-1} + \epsilon_t,
\end{equation}
where $\epsilon_t$ denotes an independently sampled random noise term at each time step, drawn from the normal distribution $\mathcal{N}(0, 2)$.

For comparative analysis, we executed Kalman filtering by (1) closed-form equation set \eqref{eq:KF-closed-form}, (2) standard sum-product message passing, and (3) spike-based message passing with LIF-SNN nodes. The results comparing the first three steps and the tenth step are summarized in Table \ref{tab:kalamn-result}. We conclude that spike-based message passing aligns closely with standard message passing and closed-form Kalman filtering.

\subsection{Bayesian Linear Regression}\label{sec:BLR-example}

Consider a given data set $\{(x_i, y_i)\}_{i=1}^N$, where $x_i \in \mathbb{R}^M$ 
and $y_i \in \mathbb{R}$. We will model this data set by a Bayesian Linear Regression (BLR) model 
\begin{subequations}\label{eq:lr-model}
 \begin{align}
p(y_i | x_i,w) &= \mathcal{N}(y_i \,|\, w^T x_i, \sigma^2) \\
p(w) &= \mathcal{N}(w \,|\, m_w, \alpha^{-1} I) \,,
\end{align}   
\end{subequations}
where $w \in \mathbb{R}^M$ holds the regression coefficients. The parameters $\sigma^2$, $m_w$, and $\alpha$ are assumed to be given. Suppose that we are interested in inferring the posterior distribution $p(w \given y_{1:N},x_{1:N})$.   Using Bayes’ rule, the posterior distribution over the weights remains Gaussian,
\begin{equation}
p(w \given D) = \mathcal{N}(w \given m_N, S_N),
\end{equation}
with posterior mean and covariance given by
\begin{equation}
S_N = (\alpha I + \beta X^\top X)^{-1},
\qquad
m_N = \beta S_N X^\top y.
\label{eq:CBLR}
\end{equation}
Where $X \in \mathbb{R}^{N \times M}$ denote the design matrix with rows $x_i^\top$,
$y = (y_1,\dots,y_N)^\top$, and $\beta = 1/\sigma^2$. See \cite[Section 3.3]{bishop2006pattern}. A mean-field approximation further simplifies the posterior by keeping the exact posterior mean but replacing the full covariance by its diagonal,
\begin{equation}
q(w) = \mathcal{N}\bigl(w \mid m_N,\; \mathrm{diag}(S_N)).
\end{equation}
This approximation assumes that the bias $w_0$ and the slope $w_1$ are independent.
According to \eqref{eq:CBLR}, classical Bayesian regression suffers from significant computational limitations because it requires large matrix multiplications and the inversion of high-dimensional matrices, both of which scale poorly as the dataset grows. In addition, when new data arrive, the relevant matrices must be recomputed or updated, leading to repeated expensive matrix operations that make the method impractical for large-scale or streaming applications.

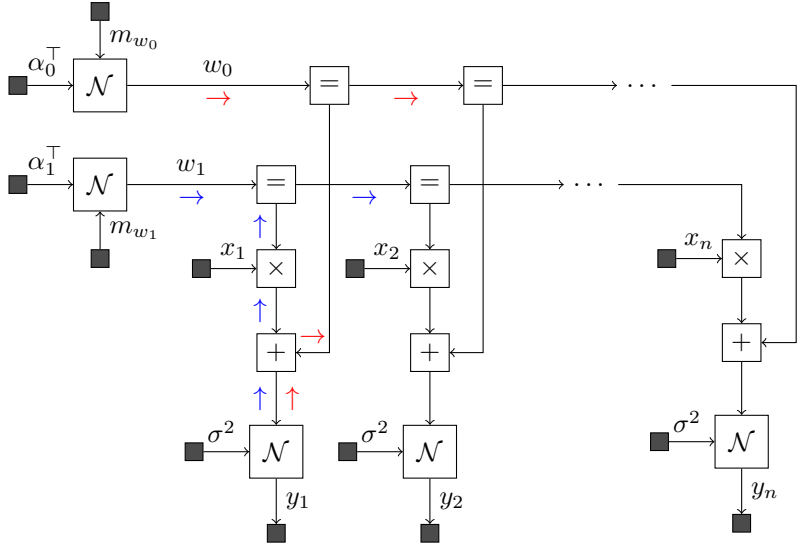
\begin{figure}[bth]
\centering
\begin{tikzpicture}
    %time step 1
    \node[box] (w) {$\mathcal{N}$};
    \node[box,  above = 6 mm of w] (b) {$\mathcal{N}$};
    \node[smallbox, right = 17 mm of w] (eq1) {$=$};
    \node[smallbox, right = 24 mm of b] (eb1) {$=$};
    \node[smallbox, below = 6 mm of eq1] (mult1) {$\times$};
    \node[clamped, left = 6 mm of mult1] (x1) {};
    \node[smallbox, below = 6 mm of mult1] (sum1) {$+$};
    \node[box, below = 7 mm of sum1] (N1) {$\mathcal{N}$};
    \node[clamped, left = 6 mm of N1] (e1) {};
    \node[clamped, below = 6 mm of N1] (y1) {};
    \node[clamped, below = 5 mm of w] (m_w) {};
    \node[clamped, left = 6 mm of w] (v_w) {};    
    \node[clamped, above = 5 mm of b] (m_b) {};
    \node[clamped, left = 6 mm of b] (v_b) {};

    \draw[->] (b) -- (eb1)node[pos=0.5, above] {$w_0$}node[pos=0.5, below]{$\textcolor{red}{\rightarrow}$};
    % \draw[->] (eb1) |- (sum1)node[pos=0.75, above]{$\textcolor{red}{\rightarrow}$}node[pos=0.75, below]{$\textcolor{blue}{\leftarrow}$};
    \draw[->] (eb1) |- (sum1)node[pos=0.75, above]{$\textcolor{red}{\rightarrow}$};
    \draw[->] (x1) -- (mult1)node[pos=0.5, above] {$x_1$};
    \draw[->] (w) -- (eq1)node[pos=0.5, above] {$w_1$}node[pos=0.5, below]{$\textcolor{blue}{\rightarrow}$};
    % \draw[<-] (mult1) -- (eq1)node[pos=0.5, left]{$\textcolor{blue}{\uparrow}$} node[pos=0.5, right]{$\textcolor{red}{\downarrow}$};
    \draw[<-] (mult1) -- (eq1)node[pos=0.5, left]{$\textcolor{blue}{\uparrow}$};
    % \draw[<-] (sum1) -- (mult1)node[pos=0.5, left]{$\textcolor{blue}{\uparrow}$}node[pos=0.5, right]{$\textcolor{red}{\downarrow}$};
    \draw[<-] (sum1) -- (mult1)node[pos=0.5, left]{$\textcolor{blue}{\uparrow}$};
    \draw[<-] (N1) -- (sum1)node[pos=0.5, left] {$\textcolor{blue}{\uparrow}$}node[pos=0.5, right]{$\textcolor{red}{\uparrow}$};
    \draw[<-] (N1) -- (e1)node[pos=0.6, above] {$\sigma^2$};
    \draw[->] (N1) -- (y1)node[pos=0.5, right] {$y_1$};
    \draw[->] (v_w) -- (w)node[pos=0.4, above] {$\alpha_1^\top$};
    \draw[->] (m_w) -- (w)node[pos=0.5, right] {$m_{w_{1}}$};
    \draw[->] (v_b) -- (b)node[pos=0.4, above] {$\alpha_0^\top$};
    \draw[->] (m_b) -- (b)node[pos=0.5, right] {$m_{w_{0}}$};

    %time step 2
    
    \node[smallbox, right = 15 mm of eq1] (eq2) {$=$};
    \node[smallbox, right = 15 mm of eb1] (eb2) {$=$};
    \node[smallbox, below = 6 mm of eq2] (mult2) {$\times$};
    \node[clamped, left = 6 mm of mult2] (x2) {};
    \node[smallbox, below = 6 mm of mult2] (sum2) {$+$};
    \node[box, below = 7 mm of sum2] (N2) {$\mathcal{N}$};
    \node[clamped, left = 6 mm of N2] (e2) {};
    \node[clamped, below = 6 mm of N2] (y2) {};

    \draw[->] (eb1) -- (eb2)node[pos=0.5, below]{$\textcolor{red}{\rightarrow}$};
    \draw[->] (eb2) |- (sum2);
    \draw[->] (x2) -- (mult2)node[pos=0.5, above] {$x_2$};
    \draw[->] (eq1) -- (eq2)node[pos=0.6, below]{$\textcolor{blue}{\rightarrow}$};
    \draw[<-] (mult2) -- (eq2);
    \draw[<-] (sum2) -- (mult2);
    \draw[<-] (N2) -- (sum2);
    \draw[<-] (N2) -- (e2)node[pos=0.6, above] {$\sigma^2$};
    \draw[->] (N2) -- (y2)node[pos=0.5, right] {$y_2$};

    % dots

    \node[right = 15 mm of eq2] (edot) {$\dots$};
    \node[right = 15 mm of eb2] (bdot) {$\dots$};
    \draw[->] (eq2) -- (edot); 
    \draw[->] (eb2) -- (bdot);

    %time step 3
    
    \node[right = 15 mm of edot] (eqn) {};
    \node[right = 15 mm of bdot] (ebn) {};
    \node[smallbox, below = 6 mm of eqn] (multn) {$\times$};
    \node[clamped, left = 6 mm of multn] (xn) {};
    \node[smallbox, below = 6 mm of multn] (sumn) {$+$};
    \node[box, below = 7 mm of sumn] (Nn) {$\mathcal{N}$};
    \node[clamped, left = 6 mm of Nn] (en) {};
    \node[clamped, below = 6 mm of Nn] (yn) {};

    \draw[->] (bdot) -- ++(20mm,0) |- (sumn);
    \draw[->] (xn) -- (multn)node[pos=0.5, above] {$x_n$};
    \draw[->] (edot) -| (multn);
    \draw[<-] (sumn) -- (multn);
    \draw[<-] (Nn) -- (sumn);
    \draw[<-] (Nn) -- (en)node[pos=0.6, above] {$\sigma^2$};
    \draw[->] (Nn) -- (yn)node[pos=0.5, right] {$y_n$};

\end{tikzpicture}
\caption{An FFG representation of the probabilistic model for Bayesian linear regression corresponding to Equation~\eqref{eq:lr-model}. Red and blue arrows indicate message passing toward computing the posteriors of~$w_0$ and~$w_1$, respectively, based on the observation~$(x_1, y_1)$.
}
\label{fig:FFG_linear-regression}
\end{figure}

Alternatively, we can automate the inference process through message passing in a factor graph. 
The corresponding FFG of a simple BLR model is presented in Fig~\ref{fig:FFG_linear-regression}. In this figure, red arrows indicate the direction of message flow contributing to the posterior of $w_0$, and blue arrows indicate the message flow contributing to the posterior of $w_1$, both based on the observation $(x_1, y_1)$.

\begin{figure}[htbp]
    \centering
    \includegraphics[width=0.95\textwidth]{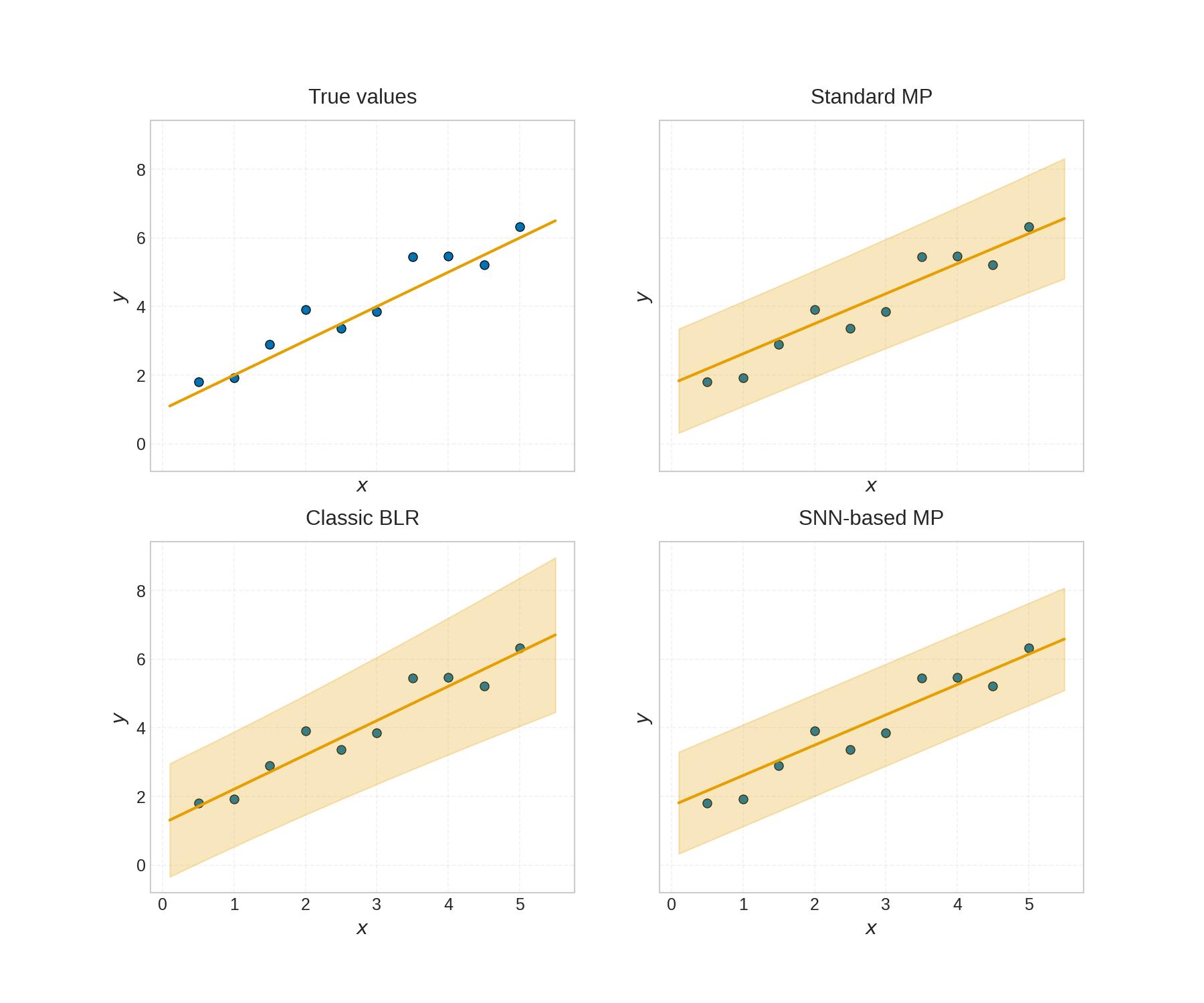}
    \caption{Comparison of linear regression results obtained using least-squares regression, classical Bayesian BLR, standard sum–product message passing, and SNN-based message passing. The plots show the inferred mean functions (solid lines) and the corresponding uncertainty bands (shaded regions), illustrating that all three probabilistic inference methods provide consistent estimates relative to the true data. }
    \label{fig:lr}
\end{figure}

For simplicity, we consider a one-dimensional case. We augment each scalar input by a bias term and write
$x_i = [1,\; x_i^{1}]^\top \in \mathbb{R}^2$ and $w = [w_0,\; w_1]^\top \in \mathbb{R}^2$. To generate a data set $D = \{(x_i,y_i)\}_{i=1}^{10}$, we randomly drew $10$ input-output pairs from the above model with a fixed weight vector $w^* = [1,1]^\top$. The training data is plotted in Fig.~\ref{fig:lr}. We compare inference performance across three approaches: classical BLR, standard message passing, and SNN-based message passing. For all methods, inference is conducted using the parameter set $\sigma^2 = 0.5$, $m_w = [0, 0]^\top$, and $\alpha = [1, 3]$. The corresponding results are shown in Fig.~\ref{fig:lr}. As illustrated, all three methods provide accurate coverage of the input data. A quantitative comparison is presented in Table~\ref{tab:linreg_results}, which shows that the proposed biologically inspired SNN-based message passing method achieves results comparable to both standard message passing and classical BLR.

\begin{table}[h!]
\centering
\setlength{\tabcolsep}{3.5pt}
\renewcommand{\arraystretch}{1.4}
\caption{Comparison of posterior weight estimates and key properties for classical BLR, standard message passing, and SNN-based message passing.}
\label{tab:linreg_results}
\begin{tabular}{|l|c|c|c|c|}
\hline
\textbf{Method} & \textbf{Posterior $w_0$} & \textbf{Posterior $w_1$} & \textbf{Scalability} & \textbf{Supports Online Updates} \\
\hline
Classic BLR        & $\mathcal{N}(1.21,\, 0.180)$  & $\mathcal{N}(0.99,\, 0.019)$   & No  & No  \\
Standard MP       & $\mathcal{N}(1.74,\, 0.075)$  & $\mathcal{N}(0.87,\, 0.006)$   & Yes & Yes \\
SNN-based MP       & $\mathcal{N}(1.72,\, 0.049)$  & $\mathcal{N}(0.88,\, 0.0001)$  & Yes & Yes \\
\hline
\end{tabular}
\end{table}

\section{Discussion}\label{sec:discussion}

A limitation of this work is the relatively slow neural encoding mechanism, which also requires explicit specification of the location parameter ranges. Future work will therefore explore alternative rate coding schemes that offer faster and more flexible representations. Additionally, the current LIF-SNN models support only message passing-based inference in linear models. Finally, we plan to investigate deployment on neuromorphic hardware to evaluate energy efficiency and assess suitability for event-based devices in practical applications.

We now examine the feasibility of event-driven message passing in greater detail. In an FFG, message updates can be triggered either by a pre-determined schedule (as in Infer.NET) or by an event-driven mechanism (as in RxInfer). An instructive analogy can be drawn between clock-driven processing in conventional CPUs and asynchronous, event-driven processing in neuromorphic systems. Traditional CPUs operate in a fixed, synchronous manner—regulated by a global clock—regardless of whether meaningful computation occurs at each cycle. By contrast, neuromorphic architectures such as Intel’s Loihi~\cite{davies2018loihi} and the SpiNNaker platform~\cite{furber2014spinnaker} implement asynchronous processing, where computation is triggered only by discrete input events. This event-driven approach offers substantially lower power consumption and greater computational efficiency, particularly when processing sparse and temporally structured data.

RxInfer’s reactive message passing process simulates a neuromorphic, event-driven computation model~\cite{bagaev2023reactive}. By adopting this paradigm, reactive message passing might provide a natural bridge between mathematical inference algorithms and biological systems such as the human brain.

\section{Conclusions}

In this work, we proposed and validated a framework for spike-based message passing in leaky integrate-and-fire spiking neural networks, enabling Gaussian belief propagation to be carried out in a biologically plausible manner. By encoding Gaussian messages into spike trains using population coding, and by designing SNN-based implementations of equality, summation, and multiplication nodes, we demonstrated that core message-passing operations can be faithfully reproduced in spiking systems.

Our experimental validation on Bayesian linear regression and Kalman filtering confirmed that SNN-based inference closely tracks standard message passing, showing that biologically inspired networks can implement principled probabilistic reasoning in both static and dynamic settings. This work, therefore, provides a constructive step toward bridging the gap between abstract Bayesian inference algorithms and neural computation.

Looking forward, several important directions emerge. First, more efficient neural coding schemes could accelerate inference and reduce reliance on predefined parameter ranges. Second, extending the framework to higher-dimensional parameter spaces would unlock more realistic applications. Third, deploying spike-based message passing on neuromorphic hardware offers the prospect of ultra-low-power, event-driven probabilistic inference, potentially aligning algorithmic neuroscience, machine learning, and neuromorphic engineering under a common framework.

In sum, this study shows that Bayesian message passing and SNNs are not only conceptually aligned but also practically integrable, providing a promising foundation for future research on brain-like, energy-efficient inference systems.

\section*{Acknowledgments}
We gratefully acknowledge funding for this study by the Eindhoven Artificial Intelligence Systems Institute (EAISI) at Eindhoven University of Technology.

% \bibliography{content/references}


\newpage
\appendix
\end{document}